\RequirePackage{fix-cm}
\documentclass[twocolumn, natbib]{svjour3}          

\smartqed  
\usepackage{graphicx}
\usepackage{mathptmx}      
\usepackage{hyperref}
\usepackage{url}
\usepackage{amssymb}
\usepackage{amsmath}

\usepackage{amsthm}

\usepackage{graphicx}
\usepackage{commath}
\usepackage{breqn}
\usepackage{tikz}
\usepackage{pgfplots}
\usepackage{caption}
\usepackage{wrapfig}
\usepackage[utf8]{inputenc} 
\usepackage[T1]{fontenc}    
\usepackage{hyperref}       
\usepackage{url}            
\usepackage{booktabs}       
\usepackage{amsfonts}       
\usepackage{nicefrac}       
\usepackage{microtype}      
\usepackage{mathtools}
\usepackage{cuted}
\usepackage{caption}
\usepackage{subcaption}
\usepackage{breqn}
\usepackage{sidecap}

\newcommand{\venue}[1]{{\scriptsize #1}}

\begin{document}

\title{Representation Learning on Unit Ball with 3D Roto-Translational Equivariance}



 \author{Sameera Ramasinghe  \and Salman Khan \and Nick Barnes \and Stephen Gould 
}


\institute{Sameera Ramasinghe \at
              Australian National University \\
              \email{sameera.ramasinghe@anu.edu.au}           
}


    \maketitle
    \begin{abstract}
      Convolution is an integral operation that defines how the shape of one function is modified by another function.  This powerful concept forms the basis of hierarchical feature learning in deep neural networks. Although performing convolution in Euclidean geometries is fairly straightforward, its extension to other topological spaces---such as a sphere ($\mathbb{S}^2$) or a unit ball ($\mathbb{B}^3$)---entails unique challenges. In this work, we propose a novel `\emph{volumetric convolution}' operation that can effectively model and convolve arbitrary functions in $\mathbb{B}^3$. We develop a theoretical framework for \emph{volumetric convolution} based on Zernike polynomials and efficiently implement it as a differentiable and an easily pluggable layer in deep networks. By construction, our formulation leads to the derivation of a  novel formula to measure the symmetry of a function in $\mathbb{B}^3$ around an arbitrary axis, that is useful in function analysis tasks. We demonstrate the efficacy of proposed volumetric convolution operation on one viable use case i.e., 3D object recognition.
    \end{abstract}

\section{Introduction}\label{sec:introduction}
Convolution-based deep neural networks have performed exceedingly well on 2D representation learning tasks (\cite{krizhevsky2012imagenet}, \cite{he2016deep}). The convolution layers perform parameter sharing (referred as '\emph{weight tying}') to learn repetitive features across the spatial domain while having lower computational cost by using local neuron connectivity. However, state-of-the-art convolutional networks can only work on Euclidean geometries and their extension to other topological spaces e.g., spheres, is an open research problem. Remarkably, the adaptation of convolutional networks to spherical domain can advance key application areas such as robotics, geoscience and medical imaging.

Some recent efforts have been reported in the literature that aim to extend convolutional networks to spherical signals. Initial progress was made by \cite{boomsma2017spherical}, who performed conventional planar convolution with carefully selected padding on a spherical-polar representation and its cube-sphere transformation \cite{ronchi1996cubed}. A recent pioneering contribution by \cite{cohen2018spherical} used harmonic analysis to perform efficient convolution on the surface of the sphere to achieve rotational equivariance. The aforementioned works however do not systematically consider radial information in a 3D shape and the feature representations are learned at fixed radii. Specifically, \cite{cohen2018spherical} estimated similarity between spherical surface and convolutional filter in $\mathbb{S}^2$, where the kernel moves in 3D rotation group $\mathbb{SO}(3)$.

\begin{figure}[t!]
\centering
\includegraphics[width=0.4\textwidth]{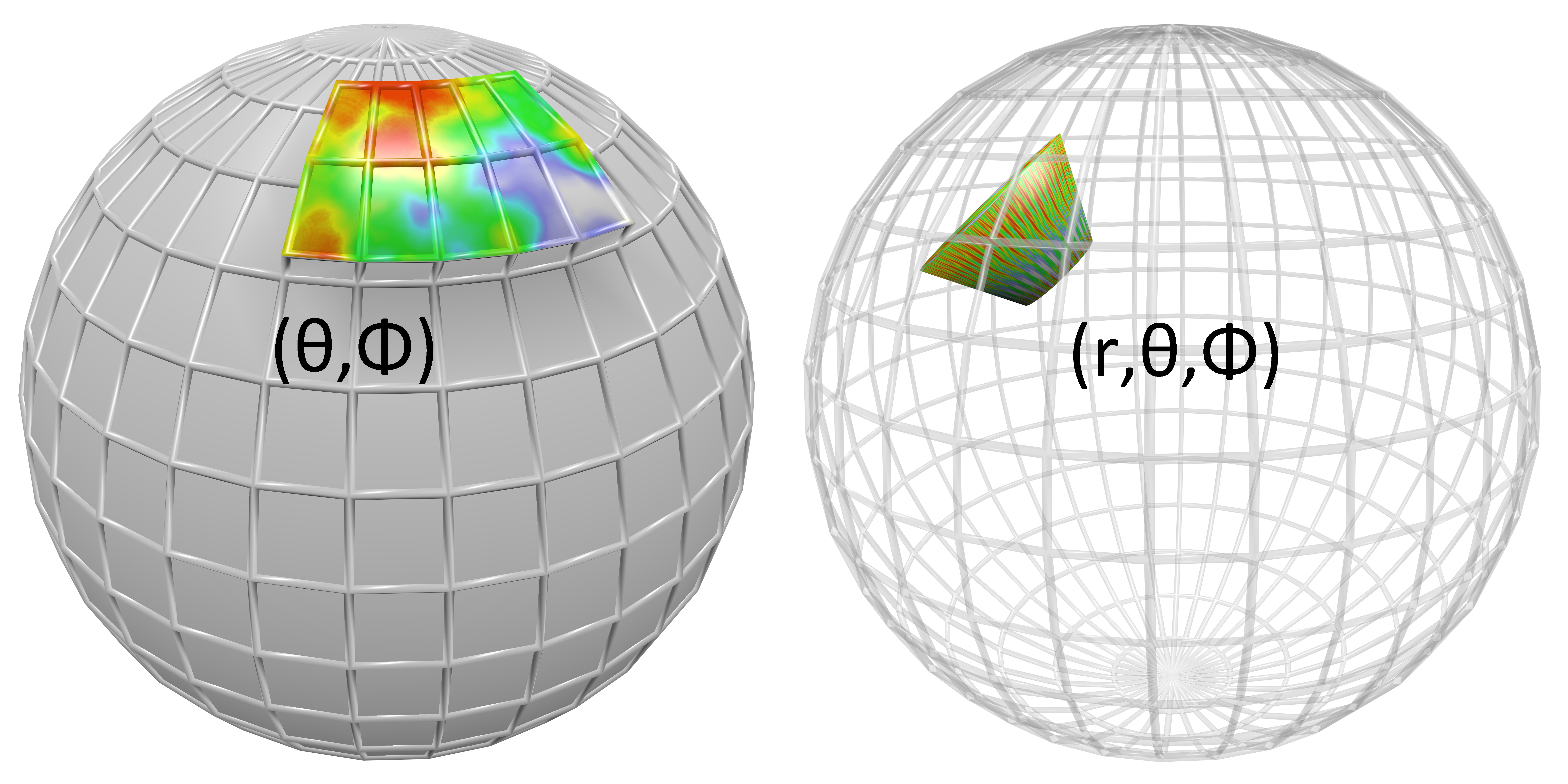}
\caption{
Fig.~1: Kernel representations of spherical convolution (\emph{left}) vs. volumetric convolution (\emph{right}). In volumetric convolution, the shape is modeled and convolved in $\mathbb{B}^3$ and in contrast, spherical convolution is performed in $\mathbb{S}^2$.
}
\label{fig:volsp}
\end{figure}

In this paper, we propose a novel approach to perform \emph{volumetric convolution} inside a unit ball ($\mathbb{B}^3$) that explicitly learns representations across the radial axis. Although we derive generic formulae to convolve functions in $\mathbb{B}^3$ we stick to two popular use cases here i.e., 3D shape recognition and retrieval. In comparison to closely related spherical convolution approaches, modeling and convolving 3D shapes in $\mathbb{B}^3$ entail key advantages: `\emph{volumetric convolution}' can capture both 2D texture and 3D shape features and can handle non-polar 3D shapes. Furthermore, volumetric convolution is equivariant to both 3D rotation and and radial translation, which enhances its ability to capture more robust features from 3D functions. We develop the theory of volumetric convolution using orthogonal Zernike polynomials \cite{canterakis19993d}, and use careful approximations to efficiently implement it as low-cost matrix multiplications. Our experimental results demonstrate significant boost over spherical convolution and confirm the high discriminative ability of features learned through volumetric convolution. Fig.~\ref{fig:volsp} compares volumetric and spherical convolution kernels.

Given that our proposed convolution operation is based on 3D orthogonal moments, we derive an explicit formula in terms of Zernike polynomials to measure the axial symmetry of a function in $\mathbb{B}^3$, around an arbitrary axis. This relation is generally applicable to function analysis tasks and here we demonstrate one particular use case with relevance to 3D shape recognition and retrieval. Specifically, we use the derived formula to propose a hand-crafted descriptor that accurately encodes the axial symmetry of a 3D shape. Moreover, we decompose the implementation of both volumetric convolution and axial symmetry measurement into differentiable steps, which enables them to be integrated in any end-to-end architecture. 


Finally, we propose an experimental architecture to demonstrate the practical usefulness of proposed volumetric convolution. A remarkable feature of our architecture is the novel spectral domain pooling layer that enhances performance, enables learning more compact features and significantly reduces the number of trainable parameters in the network. It is worth pointing out that the proposed experimental architecture is only a single possible example out of many possible architectures, and is primarily focused on demonstrating the usefulness of the volumetric convolution layer as a fully differentiable and easily pluggable layer, which can be used as a building block in end-to-end deep architectures.

The main contributions of this work include:
\begin{itemize}
\item Development of the theory for volumetric convolution that can efficiently convolve functions in $\mathbb{B}^3$ and achieve equivariance over 3D rotation and translation of local patterns.

\item Implementation of the proposed volumetric convolution as a fully differentiable module that can be plugged into any end-to-end deep learning architecture.
\item A novel formula to measure the axial symmetry of a function defined in $\mathbb{B}^3$, around an arbitrary axis using Zernike polynomials.
\item The first approach to perform volumetric convolution on 3D objects that can simultaneously model 2D (appearance) and 3D (shape) features.
\item An experimental end-to-end trainable architecture with a novel spectral pooling layer that automatically learns rich 3D shape descriptors.
\end{itemize}

The rest of the paper is structured as follows. We first introduce related work and basic concepts extensively used in the paper in Sec.~\ref{sec:rel_work} and \ref{sec:prelim} respectively followed by a detailed description of proposed volumetric convolution approach in Sec.~\ref{sec:volconv}. The axial symmetry measurement formula is derived in Sec.~\ref{sec:symmetry}. We present an example CNN architecture based on proposed convolution technique in Sec.~\ref{sec:casestudy}. In Sec.~\ref{sec:experiments} we demonstrate the effectiveness of the derived operators through extensive experiments.
Finally, we conclude the paper in Sec.~\ref{sec:conclusion}.

\section{Related Work}\label{sec:rel_work}

\textbf{Equivariance in 3D:} The convolution operation in 2D provides translation equivariance i.e., $f(t(\cdot)) = t(f(\cdot))$ where $f, t$ denote the convolution and transformation functions respectively. However, conventional convolution does not guarantee equivariance to an object's pose (rotation, translation). This is a highly desirable property in 3D shape analysis, e.g., a simple rotation of an object should not alter its category. To resolve this, Cohen and Welling \cite{cohen2018spherical} proposed Spherical CNN that performs cross-correlation after projecting images on the surface of the sphere.  \cite{WorrallG18} introduced an operator for voxelized inputs that is linearly equivariant to 3D rotations and translations. Another interesting extension of \cite{cohen2018spherical} has recently been reported in \cite{kondor2018clebsch} where Clebsch-Gordon transform is used as a spectral domain non-linearity to realize a fully Fourier domain Spherical CNN. \cite{thomas2018tensor} proposed a tensor field network that uses spherical harmonics similar to \cite{cohen2018spherical,worrall2017harmonic, WorrallG18} and exhibits local equivariance to rotations, translations and 3D point permutations. These efforts are focused on spherical projections \cite{cohen2018spherical, kondor2018clebsch} or point-clouds \cite{kondor2018n, thomas2018tensor} and cannot be directly applied to volumetric inputs. Furthermore, \cite{weiler20183d} recently proposed a solution to the problem of SE(3) equivariance by modeling 3D data as dense vector fields in 3D Euclidean space. In this work however, we focus on $\mathbb{B}^3$ to achieve radial translational and rotational equivariances over local patterns.

\textbf{Orthogonal Moments:} Orthogonal moments are useful tools for analyzing structured data. Generally, the goal of orthogonal moments is to obtain a descriptor from a data representation, that is invariant to certain deformations and transformations such as translation, rotation and scaling (TRS). Compared to geometric moments, orthogonal moments behave favorably under aforementioned transformations and therefore have been extensively used in 2D data analysis in past~\cite{hu1962visual,lin1987classification,arbter1990application, tieng1995application, khalil2001dyadic, suk1996vertex} . Many 3D TRS invariant moments are extensions of their 2D counter-parts, although extending invariant moments from 2D to 3D is not a straight forward task as rotation in 3D is not commutative. Despite this complexity, many attempts to obtain TRS invariant orthogonal moments for 3D data have been reported in literature \cite{guo1993three, reiss1992features, canterakis1996complete, canterakis19993d, flusser2003moment}. The behaviour of orthogonal moments are strongly dependant on the Hilbert space in which they are defined. For example, some moments are orthogonal inside a cube and other moments are orthogonal on a sphere or inside a  unit ball. The moments defined inside a cube are less convenient for extracting rotation invariants, compared to a sphere and a ball. Although \cite{el2017radial} and \cite{yang20153d}
proposed orthogonal moments inside unit ball, they lack two key properties, which prevents them from being used as basis functions for convolution operations: \textit{1)} loss of orthogonality under 3D rotation, \textit{2)} the \textit{completeness} of basis polynomials has not been proved in unit ball, which hampers its ability to represent an arbitrary complex function with minimal number of terms. In contrast, 3D Zernike polynomials \cite{canterakis19993d} have both aforementioned properties, which makes them an attractive choice for basis polynomials of our volumetric convolution. Recently, \cite{janssen2018design} also used generalized 3D Zernike basis functions to represent a 3D version of cake-wavelets, which then obtain orientation scores between elongated 3D structures. \emph{First}, they implement the 3D cake-wavelet functions using a discrete Fourier transform based method, which does not have an analytical description in the spatial domain. Therefore, they present an analytical version of the same 3D cake-wavelets using a 3D Zernike basis functions, followed by a continuous Fourier transform. They primarily evaluate their method on obtaining orientation scores between 3D biomedical data, such as 3D rotational Xray images, which illustrates the capacity of 3D Zernike moments in representing highly non-polar and textured data. Our work, however, is not limited to obtaining handcrafted  analytical features, as we learn deep features using the properties of 3D Zernike moments.

\textbf{3D Shape Recognition and Retrieval:} 
As a case study, this paper considers popular 3D shape recognition and retrieval problems that can directly benefit from discriminative volumetric representations. Traditionally, a diverse set of approaches have been developed for this task including handcrafted features \cite{vranic2002description, guo2016comprehensive}, unsupervised learning \cite{wu20153d, wu2016learning,khan2018adversarial} and deep learning \cite{qi2017pointnet, li2016fpnn,qi2016volumetric}. Among hand-crafted shape descriptors, a popular choice is spherical harmonics that are computed using Fourier domain coefficients \cite{vranic2002description, canterakis1996complete}. The hand-designed descriptors were targeted towards encoding both global (e.g., angle histograms \cite{ankerst19993d} and shape distributions \cite{osada2002shape}) and local 3D shape patterns (e.g., signature of histogram of orientations \cite{tombari2010unique} and  3D shape context \cite{frome2004recognizing}). More recently, \cite{wu20153d} introduced a convolutional deep belief network (DBN) that models the probabilistic distributions of 3D data. Since 3D convolutions are computationally expensive, \cite{li2016fpnn} proposed to approximate 3D spaces as volumetric fields. However, all of these deep learning based approaches perform convolutions in Euclidean space which is sub-optimal for 3D shapes. Hierarchical non-linear networks that operate on point-clouds were proposed in \cite{qi2017pointnet, qi2017pointnet++}. The proposed network design provides invariance to point permutations but do not achieve equivariance to 3D rotations.

\section{Preliminaries}\label{sec:prelim}

We have defined commonly used mathematical symbols in this paper, in Table~\ref{table:symbol}. Before delving into the details of proposed volumetric convolution, we briefly cover basic concepts below.

\begin{table}
  \caption{Mathematical symbols frequently used in this paper.}
  \label{table:symbol}
   \centering
  \begin{tabular}{p{1.5cm} p{6.5cm}}    
    \toprule
    \cmidrule(r){1-2}
    Symbol     & Description     \\
    \midrule
    $\mathbb{B}^3$ & Unit ball ($\mathbb{B}^3$) can be regarded as the set of points $u \in \mathbb{R}^3$ where ${\parallel} {u}{\parallel} < 1$. Any point in unit ball can be parameterized using coordinates $(\theta, \phi, r)$.\\
    \midrule
    $\mathbb{S}^2$ & Surface of the unit sphere ($\mathbb{S}^2$) can be regarded as the set of points $u \in \mathbb{R}^3$ where ${\parallel} {u}{\parallel} =  1$. Any point in $\mathbb{S}^2$ can be parameterized using coordinates $(\theta, \phi)$.\\
    \midrule
    $^\dagger$ & Complex conjugate.\\
    \midrule
    $\langle \cdot \,, \cdot \rangle$ & Let $f$ and $g$ be complex functions defined in a space $\Omega$. Then $\langle f,g \rangle = \int_{\Omega} f(X)g(X)^{\dagger}dX, X \in \Omega$ \\
    \midrule
    $\mathbb{SO}(3)$ & 3D rotation group.\\
    \midrule
    $\tau_{\alpha,\beta}$ & A rotation operation which aligns the north-pole with the axis towards $\alpha$ (azimuth) and $\theta$ (polar) angles.\\
    \midrule
    $Re$ & Real component of a complex value.\\
    \midrule
    $Imag$ & Imaginary component of a complex value.\\
    \midrule
    $R_{y}(\alpha)$ & A 3D rotation applied around $y$ axis.\\
    \midrule
    $Z_{n,l,m}(\theta, \phi, r)$ & $(n,l,m)^{th}$ order 3D Zernike polynomial.\\
    \midrule
    $\Omega_{n,l,m} (f)$ & $(n,l,m)^{th}$ order 3D Zernike moment of function $f$.\\
    \midrule
    $Y_{l,m}(\theta, \phi)$ & $(l,m)^{th}$ order spherical harmonic function.\\
    \midrule
    $D^l_{m,n}(\alpha, \beta, \gamma)$ & $(n,l,m)^{th}$ order Wigner D-matrix.\\
    \midrule
    $sym_{(\alpha,\beta)}(g)$ & Let $S$ be the set of functions in $\mathbb{B}^3$ which are symmetric around the axis towards $(\alpha, \beta)$, where $\alpha$ and $\beta$ are azimuth and polar angles respectively. Then $sym_{(\alpha,\beta)}(g)$ is the projection of a function $g \in \mathbb{B}^3$ into $S$.\\
    \bottomrule
  \end{tabular}
\end{table}

\subsection{Moments}
Moments are projections of a function $f$ onto a polynomial basis defined in a certain space. If the polynomial basis is orthogonal and complete, any arbitrary function in that space can be reconstructed using the corresponding moments. 
\\
\noindent
\textbf{Definition: }\textit{Let $\Phi(X_p)$ be a
$n$-variable polynomial basis of the
space $\Omega$. Let $p = (p_1, . . . , p_n)$ be a multi-index of non-negative integers which shows the highest power of the respective variables in $\Phi(X_p)$. Then general moment $M_p$ of $f$ is defined as}
\begin{equation}
    M_p = \int_{\Omega} \Phi(X_p)f(X)dX.
\end{equation}

\subsection{Equivariance}
 A function is said to be an equivariant map when its domain and codomain are acted on by the same symmetry group, and when the function commutes with the action of the group. That is, applying a symmetry transformation and then computing the function produces the same result as computing the function and then applying the transformation. We formally define equivariance as follows:

\noindent
\textbf{Definition:}\textit{ Consider a set of transformations $G$, where individual transformations are indexed as $g \in G$. Consider also a function or feature map $\phi: X \longrightarrow Y$ mapping inputs $x \in X$ to outputs $y \in Y$. Transformations can be applied to any $x \in X$ using the operator $T_g^X: X \longrightarrow X$, so that $x\longrightarrow T_g^X[x]$. The same can be done for the outputs with $y\longrightarrow T_g^Y[y]$. We say that $\phi$ is equivariant to $G$ if }
\begin{equation}
    \phi (T_g^X[x]) = T_g^Y[\phi(x)].
\end{equation}

\subsection{Spherical Harmonics}

Spherical harmonics are a set of complete and orthogonal functions defined on the surface of the unit sphere as
\begin{equation}
Y_{l,m} (\theta, \phi) = (-1)^m\sqrt{\frac{2l+1}{4\pi}\frac{(l-m)!}{(l+m)!}}P_l^m(\cos\phi)e^{im\theta},
\end{equation}
where $\phi \in [0,\pi]$ is the polar angle, $\theta \in [0, 2\pi ]$ is the azimuth angle, $l \in \mathbb{Z}^{+}$ is a non-negative integer, $m  \in \mathbb{Z}$ is an integer, $|m| < l$, and $P_l^m(\cdot)$ is the associated Legendre function
\begin{equation}
P_l^m(x) = (-1)^m \frac{(1-x^2)^{m/2}}{2^ll!}\frac{d^{l+m}}{dx^{l+m}}(x^2-1)^l.
\end{equation}
Since spherical harmonics hold the orthogonality property
\begin{equation}
    \int_{0}^{2\pi} \int_{0}^{\pi} Y_l^m (\theta, \phi) Y_{l'}^{m'}(\theta, \phi)^\dagger \sin{\phi}\, d\phi d\theta  = \delta_{l,l'} \delta_{m,m'},
\end{equation}
where $\delta_{m,m'}$ is the Kronecker delta function defined as
\begin{equation}
\delta_{m,m'} =  
\begin{cases}
    1,& \text{if } m = m' \\
    0,              & \text{otherwise}.
\end{cases}
\end{equation}

Spherical harmonics form the basis for any continuous function over $\mathbb{S}^2$ with finite energy. Therefore, a function $f:\mathbb{S}^2 \rightarrow \mathbb{R}$ can be rewritten using spherical harmonics as
\begin{equation}
f(\theta,\phi) = \sum_{l}\sum_{m=-l}^{l}\hat{f}(l,m)Y_{l,m}(\theta,\phi), \quad
\end{equation}
where $\hat{f}(l,m)$ can be obtained using
\begin{equation}
    \hat{f}(l,m) = \int_0^{\pi} \int_0^{2 \pi}f(\theta, \phi) Y_{l}^{m}(\theta, \phi)^\dagger \sin\phi \, d\phi d\theta .
\end{equation}

\subsection{Spherical Convolution}
Let $f$ and $g$ be the shape functions of the object and kernel respectively. Then $f$ and $g$ can be expressed as
\begin{equation}
f(\theta,\phi) = \sum_{l}\sum_{m=-l}^{l}\hat{f}(l,m)Y_{l,m}(\theta,\phi) \quad
\end{equation}

\begin{equation}
g(\theta,\phi) = \sum_{l}\sum_{m=-l}^{l}\hat{g}(l,m)Y_{l,m}(\theta,\phi),
\end{equation}
where $Y_{l,m}$ is the $(l,m)^{th}$ spherical harmonics function and $\hat{f}(l,m)$ and $\hat{g}(l,m)$ are $(l,m)^{th}$ frequency components of $f$ and $g$ respectively. Then, the frequency components of convolution $f*g$ can be easily calculated as
\begin{equation}
\widehat{f*g}(l,m) = \sqrt{\frac{4\pi}{2l + 1}}\hat{f}(l,m)\hat{g}(l,0)^\dagger,
\end{equation}
where $^\dagger$ denotes the complex conjugate. 

\subsection{3D Zernike Polynomials}
\label{sec:zernike}
3D Zernike polynomials are a complete and orthogonal set of basis functions in $\mathbb{B}^3$, that exhibits a \textit{form invariance} property under 3D rotation. A $(n,l,m)^{th}$ order 3D Zernike basis function is defined as
\begin{equation}
Z_{n,l,m}(r, \theta, \phi) = R_{n,l}(r)Y_{l,m}(\theta, \phi),
\end{equation}
where $R_{n,l}$ is the 3D Zernike radial polynomial defined as
\begin{equation}
    R_{n,l}(r) = \sum_{v=0}^{(n-1)/2} q^v_{nl} r^{2v +l}
\end{equation}
and  $q^v_{nl}$ is a scaler defined as
\begin{equation}
    q^v_{nl} = \frac{(-1)^{\frac{(n-l)}{2}}}{2^{(n-l)}} \sqrt{\frac{2n+3}{3}}{(n-l)\choose \frac{(n-l)}{2}}(-1)^v \frac{{\frac{(n-l)}{2}\choose v} {2(\frac{(n-l)}{2} + l + v) + 1\choose (n-l)  } }{{ \frac{(n-l)}{2} + l + v \choose \frac{(n-l)}{2}}}.
\end{equation}


$Y_{l,m}(\theta, \phi)$ is the spherical harmonics function, $n \in \mathbb{Z}^{+}$, $l \in [0, n]$, $m \in [-l, l]$ and $n-l$ is even. Since 3D Zernike polynomials are orthogonal and complete in $\mathbb{B}^3$, an arbitrary function $f(r, \theta, \phi)$ in  $\mathbb{B}^3$  can be approximated using Zernike polynomials as follows:
\begin{equation}
\label{reconstruction}
f(\theta, \phi, r) = \sum\limits_{n=0}^{\infty} \sum\limits_{l = 0}^{n} \sum\limits_{m = -l}^{l} \Omega_{n,l,m}(f) Z_{n,l,m}(\theta, \phi, r)
\end{equation}
where $\Omega_{n,l,m}(f)$ can be obtained using
\begin{equation}
\label{omega}
\Omega_{n,l,m}(f) = \int_{0}^{1} \int_{0}^{2 \pi} \int_{0}^{\pi}  f(\theta, \phi, r) {Z}^{\dagger}_{n,l,m} r^2 \sin\phi \, drd\phi d\theta.
\end{equation}
Furthermore, 3D Zernike polynomials hold the orthogonality property as follows:
\begin{equation}
\begin{split}
    \int_{0}^{1} \int_{0}^{2 \pi} \int_{0}^{\pi}  Z_{n,l,m}(\theta, \phi, r) & {Z}^{\dagger}_{n',l',m'} r^2  \sin\phi \, drd\phi d\theta \\
    & = \frac{4\pi}{3} \delta_{n,n'}\delta_{l,l'}\delta_{m,m'},
\end{split}
\end{equation}
where $\delta$ is the Kronecker delta function. In Section \ref{sec:volconv}, we will derive the proposed volumetric convolution using the concepts introduced above.


\section{Volumetric Convolution}\label{sec:volconv}

\subsection{Problem Formulation}

\begin{figure}
  
    \includegraphics[width=0.44\columnwidth]{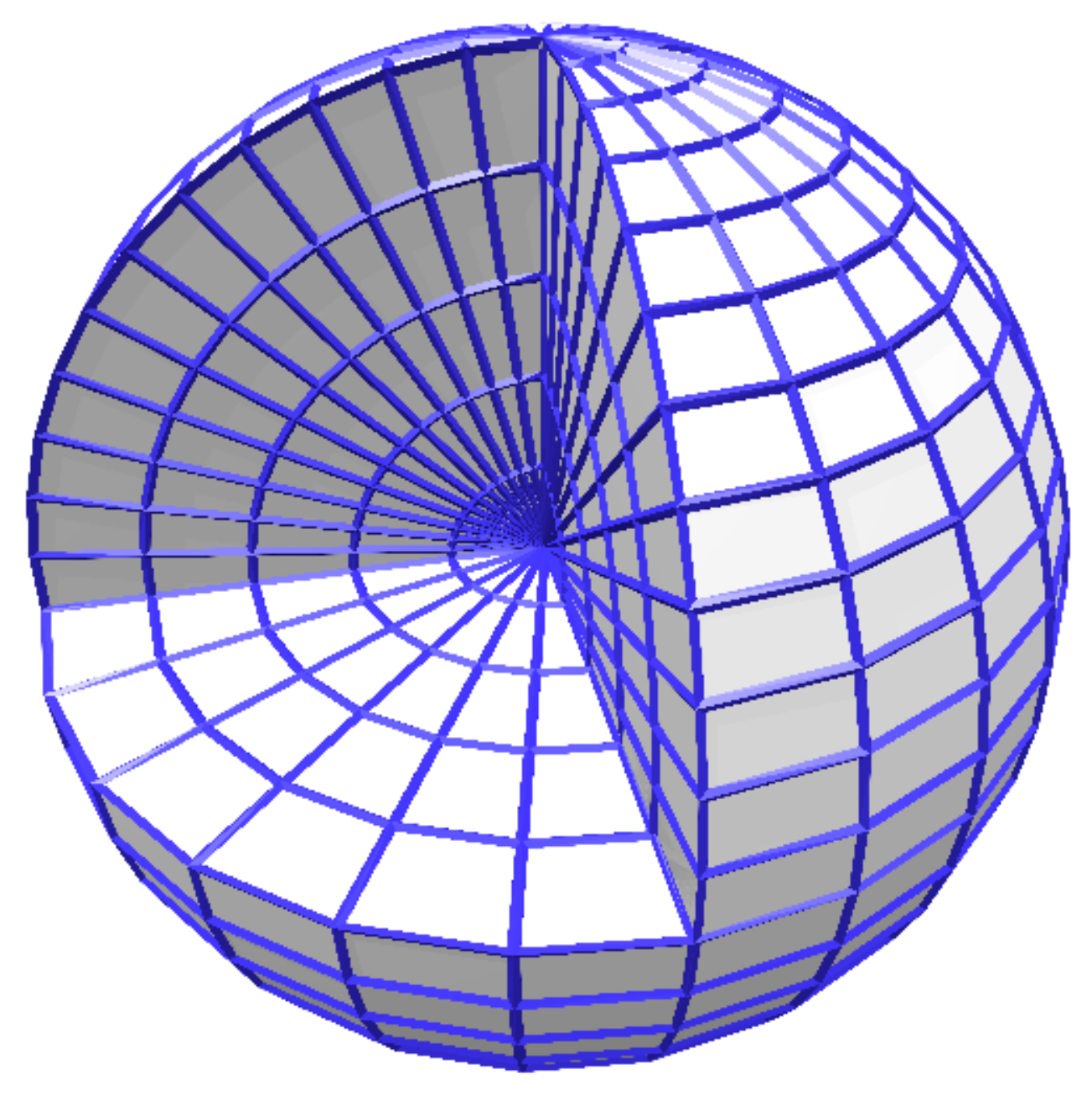}
\hfill
    \includegraphics[width=0.54\columnwidth]{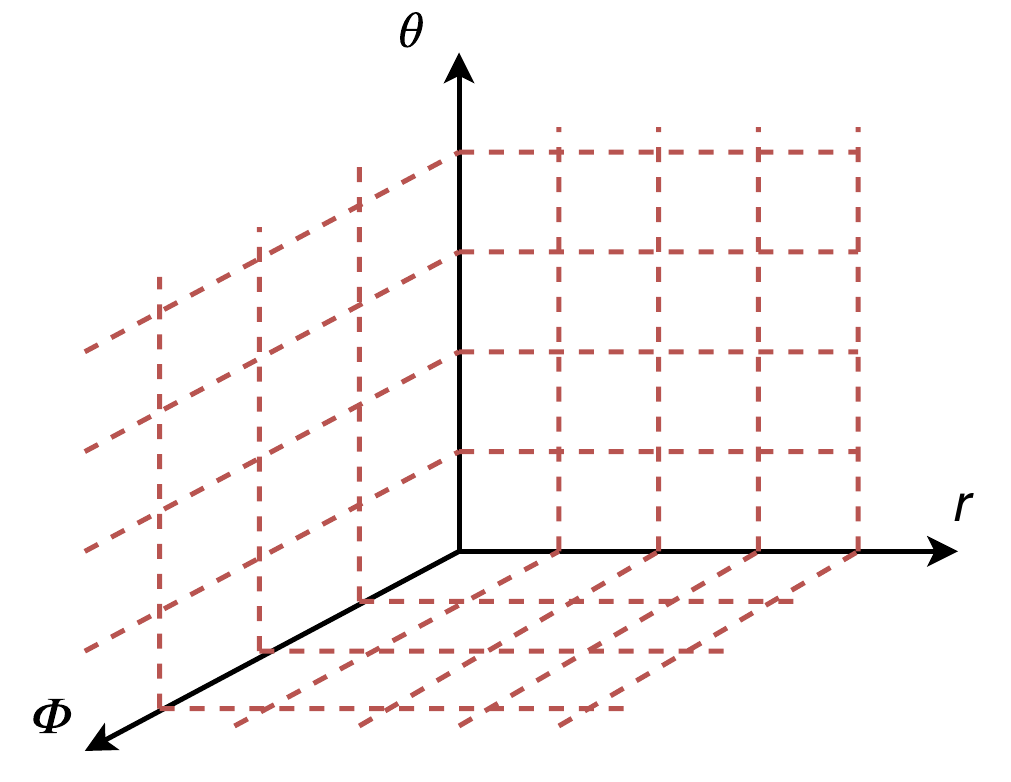}
    \caption{Grid representations in Spherical and Cartesian coordinates. \textit{Left:} The space between grid points vary with $r$ and from equator to poles. \textit{Right:} A crude approach to represent the spherical grid with a uniformly spaced grid. This approach is inaccurate as spherical grids do not have uniform spacing.}
\label{fig:grid}
\end{figure}

Convolution is an effective method to capture useful features from data represented over uniformly spaced grids points in $\mathbb{R}^n$, within each dimension of $n$. For example, gray scale images can be represented as intensities distributed over grid points in $\mathbb{R}^2$, spatio-temporal data and RGB images over grid points in $\mathbb{R}^3$, and stacked planar feature maps over grid points in $\mathbb{R}^n$. Given a shape function $f$ and a convolutional kernel $h$, this process can be more formally represented as follows:
\begin{equation}
    (f * g)(x) = \int_{\mathbb{R}^n} f(y)g(x-y)dy, \quad x,y \in \mathbb{R}^n. 
\end{equation}
 In such cases, uniformity of the grid within each dimension ensures the translation equivariance of the convolution. However, for topological spaces such as $\mathbb{S}^2$ and $\mathbb{B}^3$, it is not possible to construct such a grid due to curvilinear geometry.
This limitation is illustrated in Fig.~\ref{fig:grid}. A naive approach to perform convolution in $\mathbb{B}^3$ would be to create a uniformly spaced three dimensional grid in $(r,\theta,\phi)$ coordinates (with necessary padding) and use a regular 3D kernel. However, as shown in Fig.~\ref{fig:grid}, the spaces between adjacent points in each axis are dependant on their absolute position. Therefore, modeling such a space as a uniformly spaced grid is inaccurate. 

To overcome these limitations, we propose a novel \textit{volumetric convolution} operation which can effectively perform convolution on functions in $\mathbb{B}^3$. Volumetric convolution allows the convolution output to be a signal on $\mathbb{B}^3$, which opens up the possibility of achieving both rotation and radial translation equivariance with respect to the convolution operator.

\subsection{Convolution of functions in $\mathbb{B}^3$}
\label{sec:convolutionB3}

\subsubsection{Convolution as a function on $\mathbb{SO}(3)$}
Convolution in $\mathbb{B}^3$ can be achieved using two different approaches: 1) as a function on $\mathbb{S}^2$ or 2) as a function on $\mathbb{SO}(3)$. \cite{cohen2018spherical} showed that for functions in $\mathbb{S}^2$, modeling convolution as a function on $\mathbb{SO}(3)$ improves the capacity of the network. However for functions in $\mathbb{B}^3$, following the same approach is hampered by implementation difficulties. More precisely, if modeled as a function on $\mathbb{SO}(3)$, for all $f,g \in \mathbb{B}^3$,

\begin{equation}
    f * g (\alpha, \beta, \gamma)= \sum_{n}\sum_{l,m,m'}\hat{f}_{n,l,m}\hat{g}_{n,l,m'}D^l_{m,m'}(\alpha, \beta, \gamma),
\end{equation}
where $D$ is the Wigner D-matrix and $\alpha, \beta, \gamma$ are Euler angles. This relationship cannot be implemented as a matrix/tensor operation, since corresponding frequency components have to be extracted from spectral distributions and multiplied element-wise. Therefore, aiming for a more efficient implementation, we derive volumetric convolution as a function on $\mathbb{S}^2$, which is described in Section \ref{sec:s2conv}.


\subsubsection{Convolution as a function on $\mathbb{S}^2$}
\label{sec:s2conv}
\begin{figure}[t!]
\includegraphics[width=0.5\textwidth]{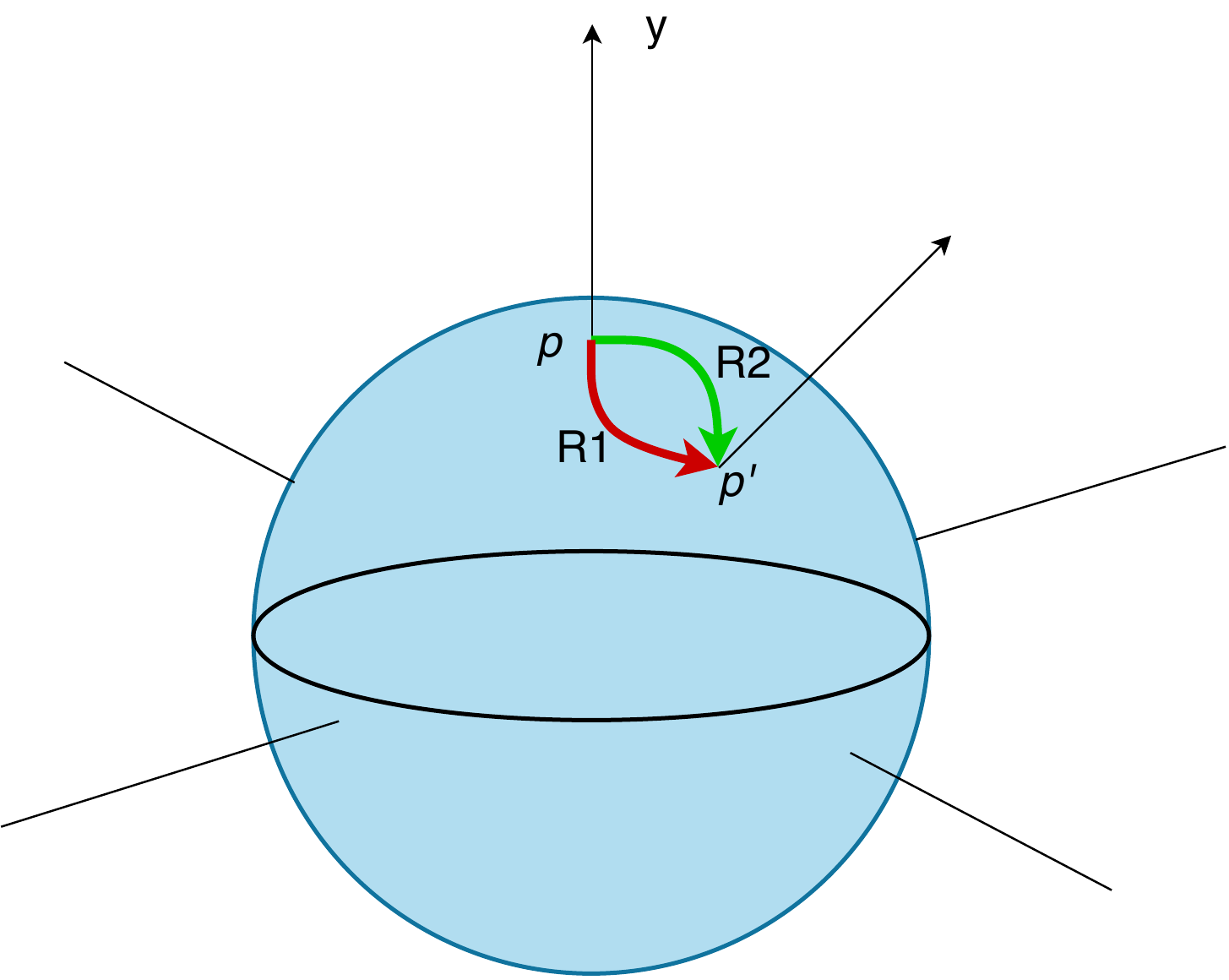}
\caption{Consider the two rotations $R_1$ and $R_2$ which takes $p$ to $p'$. Then $R_1$ and $R_2$ can be decomposed using Euler angles as $R_1 = R_y(\theta_1)R_x(\theta_2)R_y(\theta_3)$ and $R_2 = R_y(\theta_1)R_x(\theta_2)R_y(\theta_4)$, where the initial rotation around north pole is different in the two cases. Therefore, if the function is symmetric around north pole, the rotated function would only depend on $p'$.}
\label{fig:rotation}
\end{figure}

When performing convolution in $\mathbb{B}^3$ as a function on $\mathbb{S}^2$, a critical problem 
is that several rotation operations exist for mapping a point $p$ to a particular point $p'$. For example, using Euler angles, we can decompose a rotation into three rotation operations $R(\theta, \phi) = R(\theta)_yR(\phi)_zR(\theta)_y$, and the first rotation $R(\theta)_y$ can differ while mapping $p$ to $p'$ (if $y$ is the north pole) as shown in Fig.~\ref{fig:rotation}. However, if we enforce the kernel function to be symmetric around $y$, the function of the kernel after rotation would only depend on $p$ and $p'$. This observation is important, as then we can uniquely define a 3D rotation of the kernel in terms of azimuth and polar angles.

\begin{SCfigure*}
\centering
\caption{Analogy between planar and volumetric convolutions. Top \emph{(left to right)}: 2D image, kernel and planar convolution in the Cartesian plane. Bottom \emph{(left to right)}: 3D object, 3D kernel and volumetric convolution. In planar convolution the kernel translates and inner product between the image and the kernel is computed in $(x,y)$ plane. In volumetric convolution, a 3D rotation and a radial translation are applied to the kernel and the inner product is computed between  3D function and 3D kernel over $\mathbb{B}^{3}$. This allows accurate encoding of shape and texture of 3D objects. }
\includegraphics[width=0.65\textwidth]{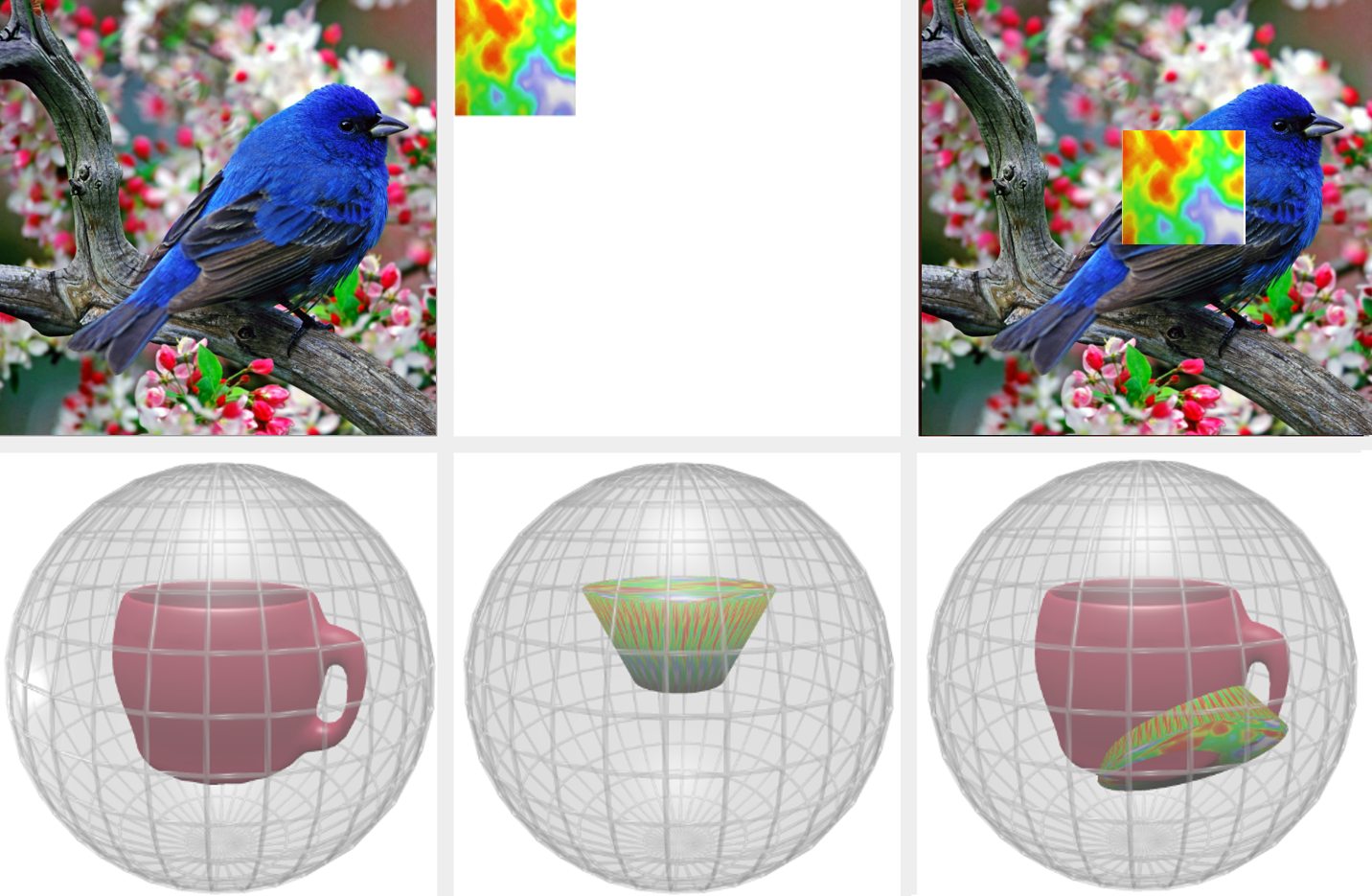}

\label{fig:analogy}
\end{SCfigure*}

Let the kernel be symmetric around $y$ and $f(\theta, \phi, r)$, $g(\theta, \phi, r)$ be the functions of object and kernel respectively. Then we define volumetric convolution as
\begin{align}
\label{conveq}
& f * g(\alpha, \beta) \coloneqq  \langle f(\theta, \phi, r), \tau_{(\alpha, \beta)}(g(\theta, \phi, r))\rangle \\ 
& = \int_{0}^1\int_{0}^{2\pi}\int_{0}^{\pi}f(\theta, \phi, r), \tau_{(\alpha, \beta)}(g(\theta, \phi, r))\sin\phi \, d\phi d\theta dr,
\end{align}
where $\tau_{(\alpha, \beta)}$ is an arbitrary rotation, that aligns the north pole with the axis towards  $(\alpha, \beta)$ direction ($\alpha$ and $\beta$ are azimuth and polar angles respectively). Eq. \ref{conveq} is able to capture more complex patterns compared to spherical convolution due to two reasons: \emph{(1)} the inner product integrates along the radius and \emph{(2)} the projection onto spherical harmonics forces the function into a polar function, that can result in information loss. It is important to note that the response of our convolution operator is a signal on $\mathbb{S}^2$, while the response of spherical convolution is a signal on 3D rotation group (\cite{cohen2018spherical}). However, we extend our convolution operator to output a function on $\mathbb{B}^3$ in Section \ref{sec:translation}, which gives multiple advantages compared to  \cite{cohen2018spherical}. Fig. \ref{fig:analogy} shows the analogy between planar convolution, spherical convolution and volumetric convolution. In Section \ref{shapemodeling} we derive formulae---preserving differentiability---to obtain 3D Zernike moments for functions in $\mathbb{B}^3$.


\subsection{Shape modeling of functions in $\mathbb{B}^3$ using 3D Zernike polynomials}
\label{shapemodeling}

Instead of using Eq. \ref{omega}, we derive an alternative method to obtain the set $\big\{\Omega_{n,l,m}\big\}$. The motivation is two fold: \emph{(1)} ease of computation and \emph{(2)} the completeness property of 3D Zernike Polynomials ensures that $\lim_{n \to \infty} \parallel f-\sum_{n}\sum_{l}\sum_{m}\Omega_{n,l,m}Z_{n,l,m}\parallel = 0$ for any arbitrary function $f$. However, since $n$ should be finite in practical implementation, aforementioned property may not hold, leading to an increased distance between the Zernike representation and the original shape. Therefore, minimizing the reconstruction error
\begin{equation}
    \sum_{(\theta,\phi,r) \in \mathbb{B}^{3}} \abs{\bar{f}(\theta, \phi, r) - f(\theta, \phi, r)},
\end{equation}
where $\bar{f}(\theta, \phi, r) = \sum_{n}\sum_{l}\sum_{m}\Omega_{n,l,m}Z_{n,l,m}$, $n\in[1,N]$ pushes the set $\big\{\Omega_{n,l,m}\big\}$ inside frequency space, where $\big\{\Omega_{n,l,m}\big\}$ has a closer resemblance to the corresponding shape. Following this conclusion, we derive the following method to obtain $\big\{\Omega_{n,l,m}\big\}$.

Since $Y_{l,m} (\theta, \phi) = (-1)^m\sqrt{\frac{2l+1}{4\pi}\frac{(l-m)!}{(l+m)!}}P_l^m(\cos\phi)e^{im\theta}$, where $P_l^m(\cdot)$ is the associated Legendre function, it can be deduced that, $Y_{l,-m} (\theta, \phi) = (-1)^m Y_{l,m}^{\dagger}(\theta, \phi)$. Using this relationship we obtain $Z_{n,l,-m} (\theta, \phi, r) = (-1)^m Z_{n,l,m}^{\dagger}(\theta, \phi, r)$ and hence approximate Eq. \ref{reconstruction} as
\begin{equation}
\label{reconstruction2}
\begin{split}
f(\theta, \phi, r) & = \sum\limits_{n=0}^{\infty} \sum\limits_{l = 0}^{n} \sum\limits_{m = 0}^{l} A_{n,l,m} Re\big\{ Z_{n,l,m}(\theta, \phi, r) \big\}  \\
& + B_{n,l,m} Img\big\{ Z_{n,l,m} (\theta, \phi, r) \big\},
\end{split}
\end{equation}
where $Re\big\{ Z_{n,l,m}(\theta, \phi, r) \big\}$ and $Img\big\{ Z_{n,l,m}(\theta, \phi, r) \big\}$ are real and imaginary components of $Z_{n,l,m}$ respectively, and $A_{n,l,m}$ and $B_{n,l,m}$ are real valued constants to be calculated. 

In practice, $f(\theta, \phi, r)$ is reconstructed using a limited number of sample points and a finite number of polynomials. Let $N$ be the order of Zernike basis functions, $K$ be the number of sample points and $\bar{f}$ be the reconstructed shape. The choice of $K$ affects both computational efficiency and the modeling accuracy. For instance, a lower value of $K$ increases the computational efficiency, but decreases modeling accuracy, and vice versa. With an appropriate choice of $K$ and $N$, using Eq.~\ref{reconstruction2}, $\bar{f}$ can be approximated in matrix form as,
\begin{equation}
\label{linear}
\bar{f} = Ua + Vb ,
\end{equation}
\noindent where,
\begin{equation}
\begin{split}
    \bar{f} = (f(\theta_1, \phi_1, r_1), \ldots ,& f(\theta_i, \phi_i, r_i), \ldots,\\
    & f(\theta_K, \phi_K, r_K))^T, (\theta_i, \phi_i, r_i) \in \mathbb{B}^3,
\end{split}
\end{equation}
\begin{equation}
    a = (A_{0,0,0}, \ldots, A_{n,l,m} , \ldots A_{N,N,N})^T,
\end{equation}
\begin{equation}
    b = (B_{0,0,0}, \ldots, B_{n,l,m} , \ldots ,  B_{N,N,N})^T,
\end{equation}
$\forall 0 \leq m \leq l \leq n\leq N$, and $n-l$ is even. $U$ and $V$ are matrices with $Re\big\{ Z_{n,l,m}(\theta_i, \phi_i, r_i) \big\}$ and $Img\big\{ Z_{n,l,m}(\theta_i, \phi_i, r_i)\big\}$ as their entries respectively, as follows:
\begin{equation}
U = 
   \begin{pmatrix} 
Re\big\{ Z_{0,0,0}(\theta_0, \phi_0, r_0) \big\} & \ldots &  Re\big\{Z_{n,l,m}(\theta_K, \phi_K, r_K)\big\}  \\
\vdots & \vdots & \vdots \\
Re\big\{ Z_{N,N,N}(\theta_0, \phi_0, r_0) \big\} & \ldots & Re\big\{ Z_{N,N,N}(\theta_K, \phi_K, r_K)\big\}  
\end{pmatrix} ,
\end{equation}

\begin{equation}
V = 
   \begin{pmatrix} 
Im\big\{ Z_{0,0,0}(\theta_0, \phi_0, r_0) \big\} & \ldots & Im\big\{Z_{n,l,m}(\theta_K, \phi_K, r_K)\big\} \\
\vdots & \vdots & \vdots \\
Im\big\{ Z_{N,N,N}(\theta_0, \phi_0, r_0) \big\} & \ldots &  Im\big\{Z_{N,N,N}(\theta_K, \phi_K, r_K)\big\} 
\end{pmatrix}.
\end{equation}
Let $X = (U,V)$ and $c = (a^T,b^T)^T$. Then, Eq. \ref{linear} can be rewritten as,
\begin{equation}
\label{eq:matform}
    \bar{f} = Xc.
\end{equation}

In other words, $c$ is the approximated set of 3D Zernike moments $\{\Omega_{n,l,m}\}$. Eq. \ref{eq:matform} can be interpreted as an overdetermined linear system,
with the set $\{\Omega_{n,l,m}\}$ as the solution. To find the least squared error solution of the Eq. \ref{eq:matform}, we use the pseudo inverse of $X$. One easy option is to use a common non-differentiable approach like singular value decomposition to find the inverse $X$. However, this imposes the condition that the inputs to the volumetric convolution layer do not depend on any learnable function. To avoid imposing this condition and allow the volumetric convolution layer to be integrated to any deep network, we propose an alternative method. \cite{li2011chebyshev}  proposed an iterative method to calculate the  pseudo inverse of a matrix. They showed that $V_{n}$ converges to $A^+$ where $A^+$ is the Moore-Penrose pseudo inverse of $A$ if
\begin{equation}
\label{equ:inverse}
V_{n+1}  = V_n(3I-AV_{n}(3I - AV_n)), n \in \mathbb{Z^+},
\end{equation}
for a suitable initial approximation $V_0$. They also showed that a suitable initial approximation would be $V_0 = \alpha A^T$ with $0 < \alpha < 2/\rho(AA^T)$, where $\rho(\cdot)$ denotes the spectral radius. Empirically, we choose $\alpha = 0.001$ in our experiments. 
Next, we derive the theory of volumetric convolution within the unit sphere.

\subsection{Convolution in $\mathbb{B}^3$ using 3D Zernike polynomials}

\label{sec: convolutiona}

We formally present our derivation of volumetric convolution using the following theorem. 

\noindent
\textbf{Theorem 1: } \textit{Suppose $f,g : \mathbb{B}^3 \longrightarrow \mathbb{R}$ are square integrable complex functions defined in $\mathbb{B}^{3}$ so that $\langle f,f \rangle < \infty$ and $\langle g,g \rangle < \infty$. Further, suppose $g$ is symmetric around north pole and $\tau (\alpha, \beta) = R_y(\alpha)R_z(\beta)$ where $R \in \mathbb{SO}(3)$. Then, }
\begin{equation}
\begin{split}
    \int_{0}^1 & \int_{0}^{2\pi}\int_{0}^{\pi}f(\theta, \phi, r), \tau_{(\alpha, \beta)}(g(\theta, \phi, r))\sin\phi\, d\phi d\theta dr \\ 
    & \equiv  \frac{4 \pi}{3} \sum\limits_{n=0}^{\infty} \sum\limits_{l = 0}^{n} \sum\limits_{m = -l}^{l} \Omega_{n,l,m}(f) \Omega_{n,l,0} (g) Y_{l,m}(\alpha, \beta), 
\end{split}
\end{equation}
\textit{where $\Omega_{n,l,m}(f), \Omega_{n,l,0}(g)$ and $Y_{l,m}(\theta, \phi)$ are $(n,l,m)^{th}$ 3D Zernike moment of $f$, $(n,l,0)^{th}$ 3D Zernike moment of $g$, and spherical harmonics function respectively.}

\noindent
\textbf{Proof: }  Completeness property of 3D Zernike Polynomials ensures that it can approximate an arbitrary function in $\mathbb{B}^3$, as shown in Eq.~\ref{reconstruction}.  
Leveraging this property, Eq. \ref{conveq} can be rewritten as
\begin{equation}
\label{linear1a}
\begin{split}
f * g(\alpha, \beta) & = \langle \sum\limits_{n=0}^{\infty} \sum\limits_{l = 0}^{n} \sum\limits_{m = -l}^{l} \Omega_{n,l,m}(f) Z_{n,l,m}, \\
& \tau_{(\alpha, \beta)}(\sum\limits_{n'=0}^{\infty} \sum\limits_{l' = 0}^{n'} \sum\limits_{m' = -l}^{l} \Omega_{n',l',m'} (g) Z_{n',l',m'}) \rangle.
\end{split}
\end{equation}
But since $g(\theta, \phi, r)$ is symmetric around $y$, the rotation around $y$ should not change the function. Which ensures
\begin{equation}
g(r, \theta, \phi) = g(r, \theta - \alpha, \phi)
\end{equation}
and hence,
\begin{equation}
\begin{split}
& \sum\limits_{n'=0}^{\infty} \sum\limits_{l' = 0}^{n'} \sum\limits_{m' = -l}^{l} \Omega_{n',l',m'} (g) R_{n',l'}(r)Y_{l',m'}(\theta, \phi) \\
& = \sum\limits_{n'=0}^{\infty} \sum\limits_{l' = 0}^{n'} \sum\limits_{m' = -l}^{l} \Omega_{n',l',m'} (g) R_{n',l'}(r)Y_{l',m'}(\theta, \phi)e^{-im'\alpha}.
\end{split}
\end{equation}
This is true, if and only if $m' = 0$. Therefore, if $g(\theta, \phi, r)$ is a symmetric function around $y$, defined inside the unit sphere, it can be rewritten as
\begin{equation}
\label{symmetrya}
\sum\limits_{n'=0}^{\infty} \sum\limits_{l' = 0}^{n'}  \Omega_{n',l',0} (g) Z_{n',l',0},
\end{equation}
which simplifies Eq. \ref{linear1a} to
\begin{equation}
\begin{split}
\label{equ1a}
f * g(\alpha, \beta) & = \langle \sum\limits_{n=0}^{\infty} \sum\limits_{l = 0}^{n} \sum\limits_{m = -l}^{l} \Omega_{n,l,m}(f) Z_{n,l,m}, \\
& \tau_{(\alpha, \beta)}(\sum\limits_{n'=0}^{\infty} \sum\limits_{l' = 0}^{n'} \Omega_{n',l',0} (g) Z_{n',l',0}) \rangle
\end{split}
\end{equation}
Using the properties of inner product, Eq. \ref{equ1a} can be rearranged as
\begin{align}
\label{propertiesa}
f * g(\alpha, \beta)  & =  \sum\limits_{n=0}^{\infty} \sum\limits_{l = 0}^{n} \sum\limits_{n'=0}^{\infty} \sum\limits_{l' = 0}^{n'} \sum\limits_{m = -l}^{l}  \Omega_{n,l,m}(f) \Omega_{n',l',0} (g) \notag\\
& \langle Z_{n,l,m}, \tau_{(\alpha, \beta)}(  Z_{n',l',0}) \rangle.
\end{align}
Consider the term $\tau_{(\alpha, \beta)}(  Z_{n',l',0})$. Then,
\begin{align}
 \tau_{(\alpha, \beta)}(  Z_{n',l',0}& (\theta, \phi, r)) = \tau_{(\alpha, \beta)}(R_{n',l'}(r)Y_{l',0}(\theta, \phi)) \notag\\
 & = R_{n',l'}(r)\tau_{(\alpha, \beta)}(Y_{l',0}(\theta, \phi)) \notag\\
 & = R_{n',l'}(r) \sum\limits_{m'' = -l'}^{l'} Y_{l',m''}(\theta, \phi)D^{l'}_{m'',0}(\alpha, \beta, \cdot),\label{eq:rot1}
\end{align}
where $D^{l}_{m,m'}$ is the Wigner D-matrix. But we know that $D^{l'}_{m'',0}(\alpha, \beta, \cdot) = Y_{l',m''}(\alpha, \beta)$. Then Eq. \ref{propertiesa} becomes
\begin{align}
\label{reduce}
f * g(\alpha, \beta) & =  \sum\limits_{n=0}^{\infty} \sum\limits_{l = 0}^{n} \sum\limits_{n'=0}^{\infty} \sum\limits_{l' = 0}^{n'} \sum\limits_{m = -l}^{l} \Omega_{n,l,m}(f) \Omega_{n',l',0} (g)\notag\\
& \sum\limits_{m'' = -l'}^{l'} Y_{l',m''}(\alpha, \beta) \langle Z_{n,l,m},  Z_{n',l',m''} \rangle ,
\end{align}
\begin{equation}
\label{equ:volconvolution}
f * g(\alpha, \beta) = \frac{4 \pi}{3} \sum\limits_{n=0}^{\infty} \sum\limits_{l = 0}^{n} \sum\limits_{m = -l}^{l} \Omega_{n,l,m}(f) \Omega_{n,l,0} (g) Y_{l,m}(\alpha, \beta), 
\end{equation}
which completes our proof. $\hfill\qed$





\subsection{Equivariance to 3D rotation group}
\label{sec:equivariance}
One key property of the proposed volumetric convolution is its equivariance to 3D rotation group. To demonstrate this we present the following theorem. 

\noindent
\textbf{Theorem 2:} \textit{Suppose $f,g : \mathbb{B}^3 \longrightarrow \mathbb{R}$ are square integrable complex functions defined in $\mathbb{B}^{3}$ such that $\langle f,f \rangle < \infty$ and $\langle g,g \rangle < \infty$. Also, let $\eta_{\alpha, \beta, \gamma}$ be a 3D rotation operator that can be decomposed into three Euler rotations $R_y(\alpha)R_z(\beta)R_y(\gamma)$ and $\tau_{\alpha, \beta}$ another rotation operator that can be decomposed into $R_y(\alpha)R_z(\beta)$. Suppose $\eta_{\alpha, \beta, \gamma}(g) = \tau_{\alpha, \beta}(g)$. Then,}
\begin{equation}
\eta_{(\alpha, \beta, \gamma)}(f) * g (\theta, \phi) = \tau_{(\alpha, \beta)}(f*g) (\theta, \phi),
\end{equation}
\textit{where $*$ is the volumetric convolution operator.}

\noindent
\textbf{Proof:} Since $\eta_{(\alpha, \beta, \gamma)} \in \mathbb{SO}(3)$, we know that $\eta_{(\alpha, \beta, \gamma)}(f(x)) = f(\eta_{(\alpha, \beta, \gamma)}^{-1}(x))$. Also we know that $\eta_{(\alpha, \beta, \gamma)} : \mathbb{R}^3 \rightarrow \mathbb{R}^3$ is an isometry.
We define,
\begin{equation}
    \langle \eta_{(\alpha, \beta, \gamma)} f, \eta_{(\alpha, \beta, \gamma)} g \rangle = \int_{B^3} f(\eta_{(\alpha, \beta, \gamma)}^{-1}(x))g(\eta_{(\alpha, \beta, \gamma)}^{-1}(x))dx.
\end{equation}

Consider the Lebesgue measure $\lambda(B^3) = \int_{B^{3}}dx$. It can be proven that a Lebesgue measure is invariant under the isometries, which gives us $dx = d\eta_{(\alpha, \beta, \gamma)} (x) = d\eta_{(\alpha, \beta, \gamma)}^{-1}(x), \forall x \in B^3$. Therefore,
\begin{equation}
\label{result}
\begin{split}
   & \langle \eta_{(\alpha, \beta, \gamma)} f, \eta_{(\alpha, \beta, \gamma)} g \rangle      =  \langle f,g \rangle \\
    & =  \int_{S^3} f(\eta_{(\alpha, \beta, \gamma)}^{-1}(x))g(\eta_{(\alpha, \beta, \gamma)}^{-1}(x))d(\eta_{(\alpha, \beta, \gamma)}^{-1}x).
\end{split}
\end{equation}

Let $f(\theta, \phi, r)$ and $g(\theta, \phi, r)$ be the object function and kernel function (symmetric around north pole) respectively. Then volumetric convolution is defined as
\begin{equation}
    f * g (\theta, \phi) =  \langle f,\tau_{(\theta, \phi)}g \rangle .
\end{equation}
Applying the rotation $\eta_{(\alpha, \beta, \gamma)}$ to $f$, we get
\begin{equation}
    \eta_{(\alpha, \beta, \gamma)}(f) * g (\theta, \phi) = \langle \eta_{(\alpha, \beta, \gamma)}(f),\tau_{(\theta, \phi)}g \rangle
\end{equation}
Using the result in Eq.~\ref{result}, we have
\begin{equation}
    \eta_{(\alpha, \beta, \gamma)}(f) * g (\theta, \phi) =  \langle f,\eta_{(\alpha, \beta, \gamma)}^{-1}(\tau_{(\theta, \phi)}g) \rangle .
\end{equation}
However, since  $\eta_{\alpha, \beta, \gamma}(g) = \tau_{\alpha, \beta}(g)$, we get
\begin{equation}
    \eta_{(\alpha, \beta, \gamma)}(f) * g (\theta, \phi) =   \langle f,\tau_{(\theta-\alpha, \phi-\beta,)}g \rangle .
\end{equation}
We know that,
\begin{equation}
\begin{split}
   &  f * g (\theta, \phi) =  \langle f,\tau_{(\theta, \phi)}g \rangle \\
    & = \sum\limits_{n=0}^{\infty} \sum\limits_{l = 0}^{n} \sum\limits_{m = -l}^{l} \Omega_{n,l,m}(f) \Omega_{n,l,0} (g) Y_{l,m}(\theta, \phi) .
\end{split}
\end{equation}
Then,
\begin{equation}
\begin{split}
    & \eta_{(\alpha, \beta, \gamma)}(f) * g (\theta, \phi) = \langle f,\tau_{(\theta-\alpha, \phi-\beta)}g \rangle \\ 
    & =  \sum\limits_{n=0}^{\infty} \sum\limits_{l = 0}^{n} \sum\limits_{m = -l}^{l} \Omega_{n,l,m}(f) \Omega_{n,l,0} (g) Y_{l,m}(\theta-\alpha, \phi-\beta)  \\
    & = (f*g)(\theta-\alpha,\phi-\beta) = \tau_{(\alpha, \beta)}(f*g) (\theta, \phi) .
\end{split}
\end{equation}

Hence, we achieve equivariance over 3D rotations. $\hfill\qed$

In simple terms, the theorem states that if a 3D rotation is applied to a function defined in $\mathbb{B}^{3}$ Hilbert space, the output feature map after volumetric convolution exhibits the same rotation. The output feature map however, is symmetric around north pole, hence the rotation can be uniquely defined in terms of azimuth and polar angles. In Section \ref{sec:symmetry} we derive the axial symmetry measure of a function in $\mathbb{B}^3$ around an arbitrary axis using 3D Zernike polynomials.

\section{Axial symmetry measure of a function in $\mathbb{B}^3$ around an arbitrary axis}
\label{sec:symmetry}
\begin{figure}[t!]
\centering
\includegraphics[width=0.5\textwidth]{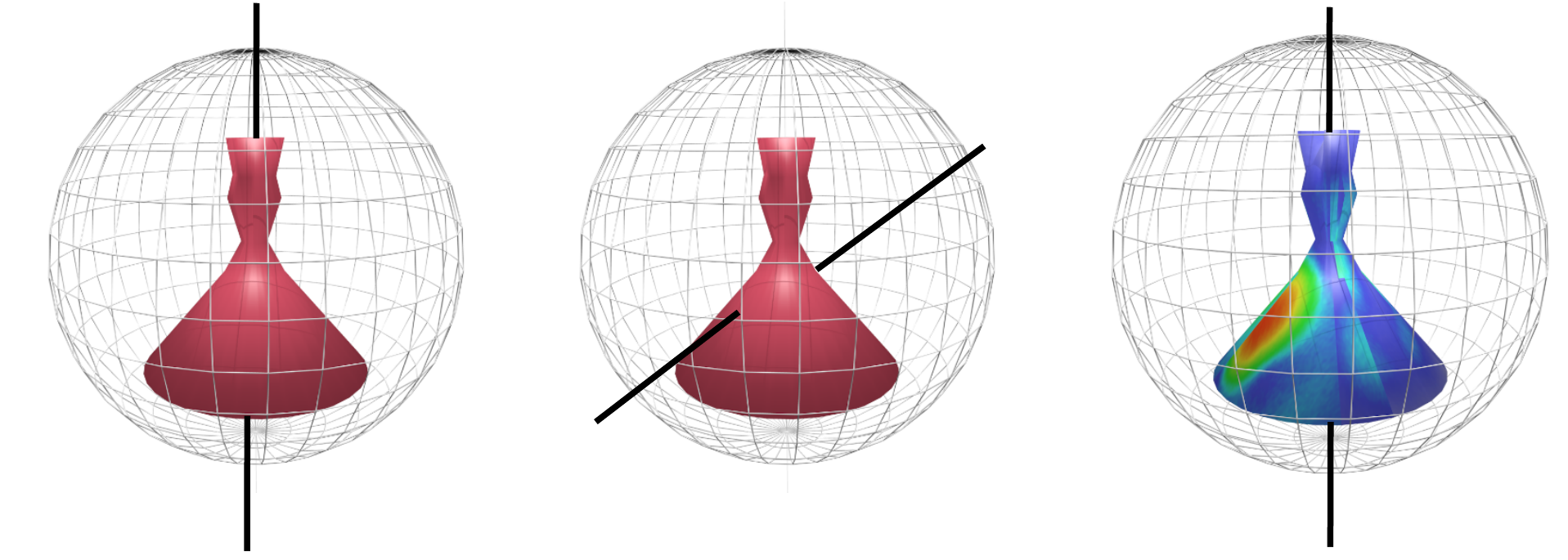}
\caption{Three cases of axial symmetry: \textit{left:} axial symmetry measurement is high, as both point values and overall shape of the function are symmetric around the axis. \textit{Middle:} axial symmetry measurement is low, as overall shape of the function is not symmetric around the axis. \textit{Right:} axial symmetry measurement is low, as point values of the function are not symmetrically distributed around the axis.}
\label{fig:symmetry}
\end{figure}

In this section we present the following proposition to obtain the axial symmetry measure of a function in $\mathbb{B}^3$ around an arbitrary axis using 3D Zernike polynomials. An illustration of axial symmetry measurement is shown in Fig.~\ref{fig:symmetry}.

\noindent
\textbf{Proposition: } \textit{Suppose $g : \mathbb{B}^3 \longrightarrow \mathbb{R}^{3}$ is a square integrable complex function defined in $\mathbb{B}^{3}$ such that $\langle g,g \rangle < \infty$. Then, the power of projection of $g$ in to $S = \{Z_i\}$ where $S$ is the set of Zernike basis functions that are symmetric around an axis towards $(\alpha, \beta)$ direction is given by}
\begin{equation}
\label{sym_finalt}
\parallel sym_{(\alpha, \beta)}\left[g (\theta, \phi, r) \right]\parallel = \sum_{n}\sum_{l=0}^{n}\parallel\sum_{m=-l}^{l}\Omega_{n,l,m}Y_{l,m}(\alpha, \beta)\parallel^2 ,
\end{equation}
\textit{where $\alpha$ and $\beta$ are azimuth and polar angles respectively}. 

\noindent
\textbf{Proof: }The subset of complex functions which are symmetric around north pole is $S = \big\{ Z_{n,l,0} \big\}$. Therefore, projection of the function into $S$ gives
\begin{equation}
sym_{y}\left[g(\theta, \phi, r)\right] = \sum_{n}\sum_{l=0}^{n}\langle f,Z_{n,l,0}\rangle Z_{n,l,0}(\theta, \phi, r).
\end{equation}

To obtain the symmetry function around any axis which is defined by $(\alpha, \beta)$, we rotate the function by $(-\alpha, -\beta)$, project into $S$, and finally compute the power of the projection
\begin{equation}
\label{syma}
sym_{(\alpha, \beta)}\left[g(\theta, \phi, r)\right] = \sum_{n,l}\langle \tau_{(-\alpha, -\beta)}(f),Z_{n,l,0}\rangle  Z_{n,l,0}(\theta, \phi, r).
\end{equation}

For any rotation operator $\tau$, and for any two points defined on a complex Hilbert space, $x$ and $y$,
\begin{equation}
\langle \tau(x), \tau(y)\rangle _H = \langle x,y \rangle_H.
\end{equation}
Applying this property to Eq.~\ref{syma} gives
\begin{equation}
sym_{(\alpha, \beta)}\left[g(\theta, \phi, r)\right] = \sum_{n,l}\langle f,\tau_{(\alpha, \beta)}(Z_{n,l,0})\rangle  Z_{n,l,0}(\theta, \phi, r).
\end{equation}
Using Eq. \ref{reconstruction} we get
\begin{align}
sym_{(\alpha, \beta)}\left[g(\theta, \phi, r)\right] = \sum_{n}\sum_{l=0}^{n} & \langle \sum_{n'} \sum_{l'=0}^{n'} \sum_{m'=-l'}^{l'} \Omega_{n'l'm'}Z_{n',l',m'},\notag\\
& \tau_{(\alpha, \beta)}(Z_{n,l,0})\rangle Z_{n,l,0}(\theta, \phi, r).
\label{sym2}
\end{align}
Using properties of inner product Eq. \ref{sym2} further simplifies to
\begin{align}
sym_{(\alpha, \beta)}\left[g(\theta, \phi, r)\right]  = & \sum_{n}\sum_{l=0}^{n}\sum_{n'} \sum_{l'=0}^{n'} \sum_{m'=-l'}^{l'} \Omega_{n'l'm'}\langle Z_{n',l',m'},\notag \\
& \tau_{(\alpha, \beta)}(Z_{n,l,0})\rangle  Z_{n,l,0}(\theta, \phi, r).
\end{align}
Using the same derivation as in Eq.~\ref{eq:rot1},
\begin{equation}
\begin{split}
sym_{ (\alpha,\beta)}& \left[g(\theta, \phi, r)\right]  =  \sum_{n}\sum_{l=0}^{n}\sum_{n'} \sum_{l'=0}^{n'} \sum_{m'=-l'}^{l'} \Omega_{n'l'm'} \\ 
& \sum_{m''=-l}^{l}Y_{l,m''}(\alpha, \beta) \langle Z_{n',l',m'}, Z_{n,l,m''} \rangle  Z_{n,l,0}(\theta, \phi, r).
\end{split}
\end{equation}
Since 3D Zernike Polynomials are orthogonal we get
\begin{equation}
\begin{split}
    sym_{(\alpha, \beta)}& \left[g (\theta, \phi, r)\right]  \\
    & = \frac{4\pi}{3}\sum_{n}\sum_{l=0}^{n}\sum_{m=-l}^{l}\Omega_{n,l,m}Y_{l,m}(\alpha, \beta) Z_{n,l,0}(\theta, \phi, r).
\end{split}
\end{equation}

In signal theory the power of a function is taken as the integral of the squared function divided by the size of its domain. Following this we get
\begin{equation}
\label{sym_f}
\begin{split}
\parallel sym_{(\alpha, \beta)}& \left[g (\theta, \phi, r) \right] \parallel \\
& =  \langle(\sum_{n}\sum_{l=0}^{n}\sum_{m=-l}^{l}\Omega_{n,l,m}Y_{l,m}(\alpha, \beta)) Z_{n,l,0}(\theta, \phi, r), \\
& (\sum_{n'}\sum_{l'=0}^{n'}\sum_{m'=-l'}^{l'}\Omega_{n',l',m'}Y_{l',m'}(\alpha, \beta) Z_{n',l',0}(\theta, \phi, r))^{\dagger}\rangle .
\end{split}
\end{equation}
We drop the constants here since they  do not depend on the frequency. Simplifying Eq. \ref{sym_f} gives
\begin{equation}
\begin{split}
\parallel sym_{(\alpha, \beta)}\left[g (\theta, \phi, r) \right]  \parallel = \sum_{n}\sum_{l=0}^{n}\sum_{m=-l}^{l}\sum_{m'=-l}^{l} & \Omega_{n,l,m}Y_{l,m}(\alpha, \beta) \\
& \Omega_{n,l,m'}Y_{l',m}(\alpha, \beta),
\end{split}
\end{equation}
which leads to
\begin{equation}
\label{sym_final}
\parallel sym_{(\alpha, \beta)} \left[ g (\theta, \phi, r) \right]   \parallel = \sum_{n}\sum_{l=0}^{n}\parallel \sum_{m=-l}^{l}\Omega_{n,l,m}Y_{l,m}(\alpha, \beta)\parallel^2.
\end{equation}
which completes our proof. $\hfill\qed$ 

Using our derivation, one can obtain the distribution of symmetry the object has around a set of axes. However, to compare two objects by the amount of symmetry it has around a specific axis, it is needed to normalize the symmetry measurement by dividing the final result with the norm of the unprojected function.


\section{A case study: Representation Learning on 3D objects}\label{sec:casestudy}

A 2D image is a function on Cartesian plane, where a unique value exists for any $(x,y)$ coordinate. Similarly, a polar 3D object can be expressed as a function on the surface of the sphere, where any direction vector $(\theta, \phi)$ has a unique value. To be precise, a 3D polar object has a boundary function in the form of $f:\mathbb{S}^2 \to [0,\infty]$.



\begin{figure}[t!]
\centering
\includegraphics[width=0.5\textwidth]{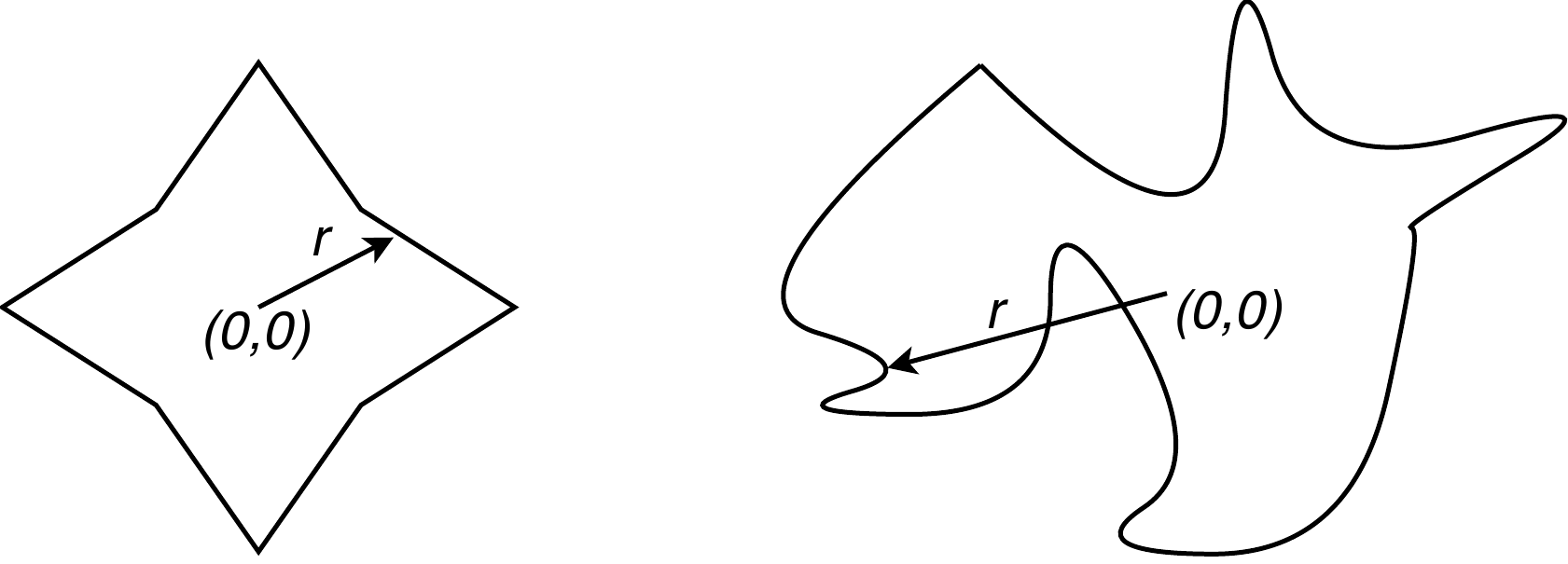}
\caption{A 2D illustration of polar and non-polar shapes.}
\label{fig:polar}
\end{figure}

Translation of the convolution kernel on $(x,y$) plane in 2D case, extends to movements on the surface of the sphere in $\mathbb{S}^2$. If both the object and the kernel have polar shapes, this task can be tackled by projecting both the kernel and the object onto spherical harmonic functions. However, using spherical convolution to capture features from 3D point clouds entail three critical limitations. \emph{First}, the projection of the points on to the surface of the sphere smoothens the overall shape in to a polar one. In other words, since it formulates the shape as a function on $(\theta. \phi)$, it restricts the representation of complex (non-polar) objects. An illustration of 2D polar and non-polar shapes is shown in Fig.~\ref{fig:polar}. \emph{Second}, the integration happens over the surface of the sphere, which is unable to capture patterns across radius. \emph{Third}, spherical convolution is equivariant to only 3D rotation group. 

These limitations can be addressed by representing and convolving the shape function inside the unit ball ($\mathbb{B}^3$). Representing the object function inside $\mathbb{B}^3$ allows the function to keep its complex shape information without any deterioration since each point is mapped to unique coordinates $(r,\theta,\phi)$, where $r$ is the radial distance, $\theta$ and $\phi$ are azimuth and polar angles respectively. Additionally, it allows encoding of 2D texture information simultaneously. The volumetric convolution can also achieve equivariance to both 3D rotation and radial translation of local patterns. Fig. \ref{fig:volsp} compares volumetric convolution and spherical convolution.

We conduct experiments on 3D objects with uniform surface values, therefore in this work we use the following transformation to apply a simple surface function to the 3D objects:
\begin{equation}
f(\theta, \phi, r)=  
\begin{cases}
    r,& \text{if surface exists at } (\theta, \phi, r)\\
    0,              & \text{otherwise}.
\end{cases}
\end{equation} 

\subsection{Equivariance to 3D radial translation}
\label{sec:translation}

Consider the case where the kernel is shifted along the radius and then convolved with the input function. Let $R_{n,l}$ be the linear component of the Zernike polynomial.  Then, if we consider only the linear component, shifting the kernel by $r'$ and then convolving with the input function gives,
\begin{align}
    \int_0^1 R_{nl}(r) R_{n'l} & (r-r') r^2 dr \notag\\
    & = \int_0^1 \sum_{v=0}^{\frac{n-l}{2}}q_{nl}^v r^{2v+l} \sum_{v'=0}^{\frac{n'-l}{2}}q_{n'l}^{v'}(r-r')^{2v'+l}r^2dr \notag\\\,
    & = \sum_{v=0}^{\frac{n-l}{2}} q_{nl}^v  \sum_{v'=0}^{\frac{n'-l}{2}}q_{n'l}^{v'} \int r^{2v+l} (r-r')^{2v'+l}r^2dr,
\end{align}
which produces the result
\begin{equation}
    \sum_{v=0}^{\frac{n-l}{2}} q_{nl}^v  \sum_{v'=0}^{\frac{n'-l}{2}}q_{n'l}^{v'} \frac{(-r')^{l+2v'} \prescript{}{2}F_1 [-l-2v',3+l+2v;4+l+2v;\frac{1}{r'}\big]}{3+l+2v},
\end{equation}
where $\prescript{}{2}F_1$ is the hypergeometric function. This complex relationship hampers achieving equivariance to 3D translation directly using properties of 3D Zernike Polynomials, preserving differentiability. Hence, we follow an alternative approach to achieve this task which is explained below.

Let's consider the input function $f(\theta_i,\phi_i,r_i), \forall (\theta_i,\phi_i,r_i) \in \mathbb{B}^3$. Then, let us define $q_k$ as,
\begin{equation}
    q_k = 0.1k, \forall 0\leq k < 10, k \in \mathbb{Z}.
\end{equation}

Next, we extract the sets of points $f'_k \in f(\theta_i,\phi_i,r_k)$, $\forall q_k < r_k < q_{k + 1}$. Then, let's consider the convolution kernel $g(\theta_i,\phi_i,r_i)$, $\forall (\theta_i,\phi_i,r_i) \in \mathbb{B}^3$. We take the radially translated kernel,
\begin{equation}
Tr_{(q_k)}\big[g(\theta_i,\phi_i,r_i)] =  g(\theta_i,\phi_i,r_i - q_k),
\end{equation}
where, $0 \leq r_i - q_k < 1$. Here, $Tr_{(q_k)}\big[\cdot]$ is radial translation by $q_k$.

\begin{figure}[t!]
\centering
\includegraphics[width=0.4\textwidth]{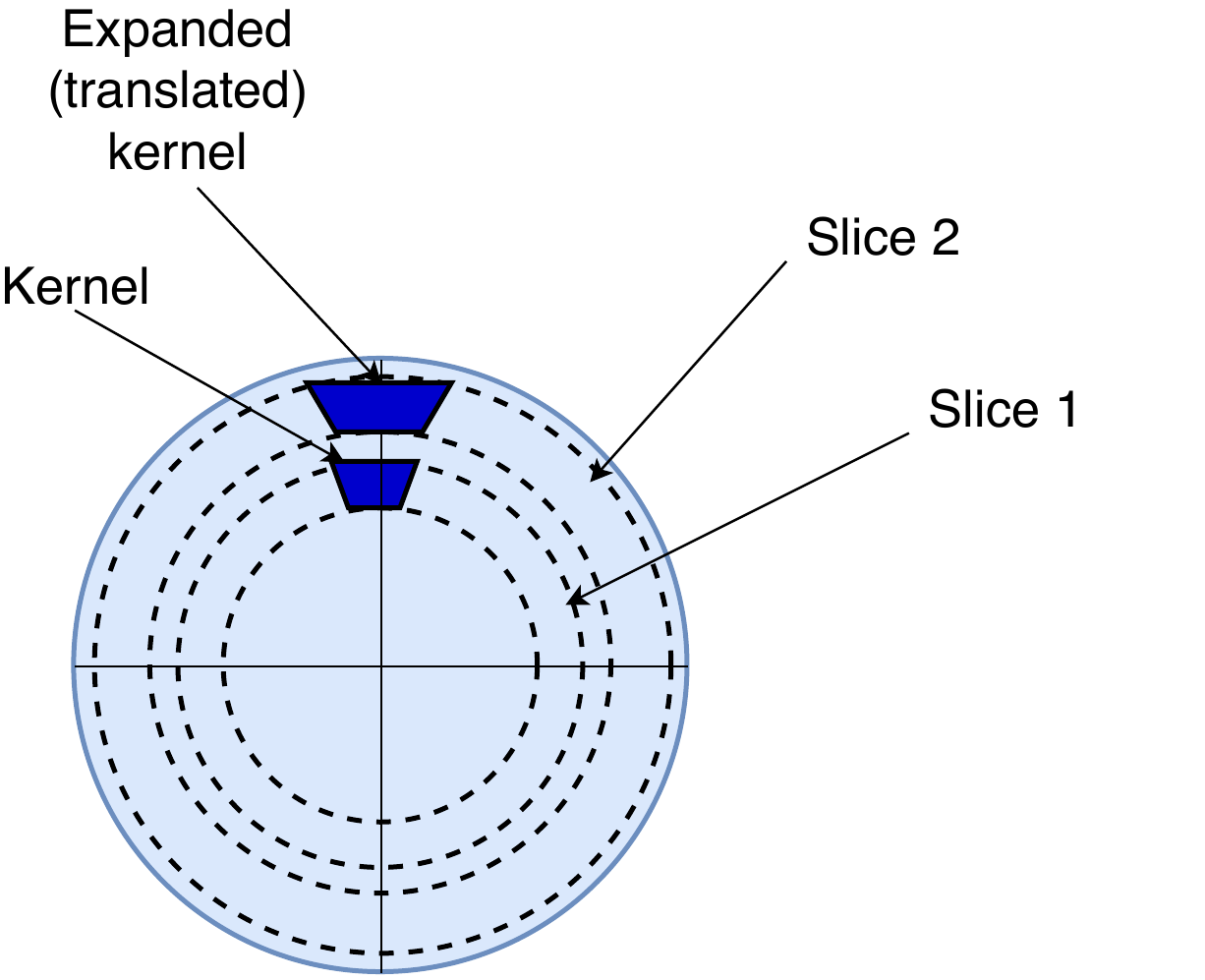}
\caption{Weight sharing across radius.}
\label{fig:weight}
\end{figure}

\begin{SCfigure*}
\centering
\caption{Illustration of volumetric convolution with weight sharing across radius. For the sake of clarity, this illustration only shows a single convolutional kernel. We bisect and show a cross section of the resultant feature map on right for better visualization. In the resultant feature map, each spherical heatmap corresponds to the response at a specific translation of the kernel. Each value in a spherical heatmap corresponds to the response at a specific 3D orientation of the kernel at a specified translation. Therefore, the resultant feature map is a signal on $\mathbb{B}^3$, which allows us to achieve equivariance over 3D rotation and radial translation of local patterns.}
\includegraphics[width=0.6\textwidth]{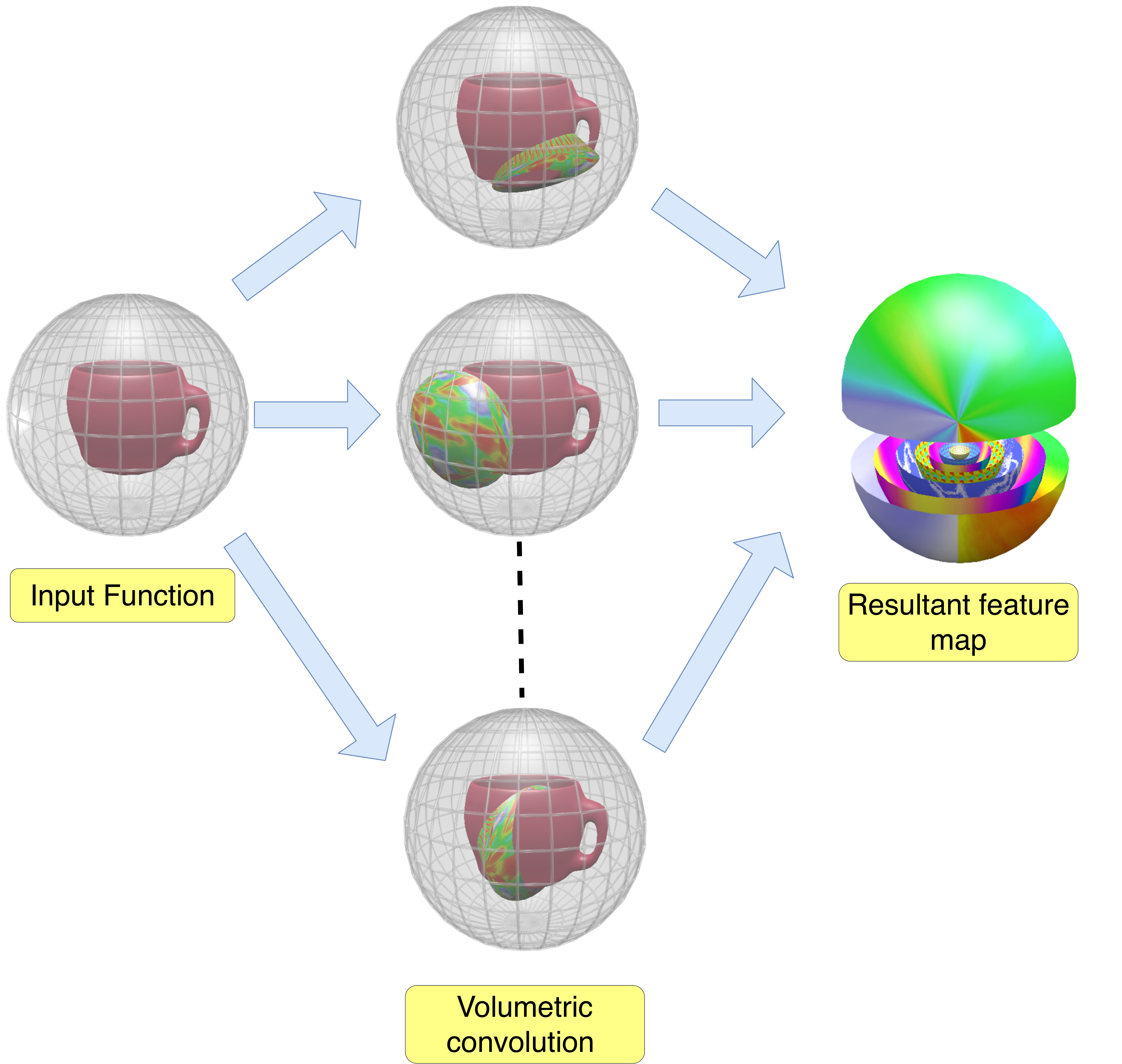}
\label{fig:toy}
\end{SCfigure*}

Finally, we perform convolution between $f'_k$ and $Tr_{(q_k)}\big[g]$ for each $k$, as graphically illustrated in Fig.~\ref{fig:weight}, which extends the response of our convolution operator to $\mathbb{B}^3$ as,
\begin{equation}
    (f'_k*g)(\alpha, \beta, q_k) = f'_k*Tr_{(q_k)}\big[\tau_{(\alpha,\beta)}g].
\end{equation}
Convolving the aforementioned point sets individually with corresponding radially translated kernel values allows us to share weights along radius, in other words, achieve equivariance over 3D radial translation for local feature patterns. Furthermore, the output of the convolution gives us a dense representation in $\mathbb{B}^3$, as illustrated in Fig. \ref{fig:toy}. Equivariance to 3D radial translation can be more formally illustrated as follows.

Let $p = f(\theta_i,\phi_i,r_i)$, $\forall (\theta_i,\phi_i,r_i) \in P$, where $P$ is a set of points which belongs to a local feature pattern of a function in $\mathbb{B}^3$. Then, we perform convolution on $p$ with a kernel $h$,
\begin{equation}
   (p * h) (\alpha, \beta, q_k) = p * Tr_{(q_k)}\big[\tau_{(\alpha,\beta)}h].
\end{equation}

\noindent Suppose we translate the local feature pattern radially. Then,

\begin{equation}
    (Tr_{r'}\big[ p] * h) (\alpha, \beta, q_k) = (p(r-r') * h(r)) (\alpha, \beta, q_k)
\end{equation}

\noindent Let $r'' = r - r'$. Then,

\begin{equation}
\begin{split}
    (Tr_{r'}\big[ p] * h) (\alpha, \beta, q_k) & =  (p(r'') * h(r'+r'')) (\alpha, \beta, q_k),\\
   & = (p(r'') * h(r'')) (\alpha, \beta, q_k-r'),\\
    & = Tr_{(r')}\big[(p(r'') * h(r'')) (\alpha, \beta, q_k)].\quad
\end{split} 
\end{equation}

Hence, we achieve equivariance over 3D radial translation of local patterns. The intuition behind the aforementioned process is that if a specific shape attribute of the object (not necessarily the whole object) translates along the radius, the corresponding output feature pattern of would also translate along the radius of the output feature map, which is in $\mathbb{B}^3$. A practical requirement to achieve this equivariance is that the kernel should cover approximately the same area as the local pattern. We achieve this requirement by designing the kernel as a concentrated set of points over a limiterd area, in the spatial domain.



\subsection{Adaptive Weighted Frequency Pooling}
\label{sec:frequency}
 Feature pooling helps in aggregating information in spatial or filter response domain. Although feature pooling is an established mechanism in spatial domain, frequency domain pooling is largely an unsolved problem. Here, we propose a simple frequency pooling approach that fuses information across different frequencies to learn more compact and discriminative features.

Let us reconsider the proposed volumetric convolution formula at a specific translation of the kernel,
\begin{equation}
\label{equ:convspec}
   f  * g(\alpha, \beta) \equiv \frac{4 \pi}{3} \sum\limits_{n=0}^{\infty} \sum\limits_{l = 0}^{n} \sum\limits_{m = -l}^{l} \Omega_{n,l,m}(f) \Omega_{n,l,0} (g) Y_{l,m}(\alpha, \beta).
\end{equation}
As evident from the above formula, the response is also in spatial domain and is a signal on $S^2$. However, any signal on $S^2$ can be completely characterized by its corresponding spherical harmonic frequencies. To leverage this property, we rearrange Equation \ref{equ:convspec} as follows,
\begin{equation}
   f  * g(\alpha, \beta) \equiv  \frac{4 \pi}{3}  \sum\limits_{l = 0}^{n} \sum\limits_{m = -l}^{l} \Bigg( \sum\limits_{n=0}^{\infty}  \Omega_{n,l,m}(f) \Omega_{n,l,0} (g) \Bigg) Y_{l,m}(\alpha, \beta).
\end{equation}

It is obvious that $S_{l,m} = \Bigg( \sum\limits_{n=0}^{\infty}  \Omega_{n,l,m}(f) \Omega_{n,l,0} (g) \Bigg)$ represents $(l,m)^{th}$ spherical harmonics frequency of the response of volumetric convolution. Since, in practice we use $n=6$, the set $\{S_{l,m}\}, \forall (m,l),$ where $ 0 \leq l \leq 5$ and $-l \leq m \leq l$, encodes all the shape information in a low dimensional vector, compared to spatial domain representation. Therefore, instead of spatial domain representation, we connect the spectral representation to the fully connected layer.

Furthermore, it can be observed that the set  $\{S_{l,m}\}$ is within the linear span of $\Omega_{n,l,m}(f) \cup \Omega_{n,l,0}(g)$. Therefore, instead of calculating  $\{S_{l,m}\}$ in a precise manner, we take the outer product between $\Omega_{n,l,m}(f)$ and $\Omega_{n,l,0}(g)$ to get a dense frequency map $\Omega$ as follows:
\begin{equation}
    \Omega = \Omega_{n,l,m}(f) (\Omega_{n,l,0}(g)^T 
\end{equation}
where  $\Omega \in \mathbb{R}^{(100 \times 100)}$ and $\Omega_{n,l,m}(f), \Omega_{n,l,0}(g) \in \mathbb{R}^{(100 \times 1)}$.  Then, we obtain two dense weighted frequency maps, $F_1 \in \mathbb{R}^{(100 \times 100)}$ and $F_2 \in \mathbb{R}^{(100 \times 100)}$, by
\begin{equation}
   F_1 = \Omega \circ W_1, \;\text{and}\;
    F_2 = \Omega \circ W_2,
\end{equation}

where $\circ$ is the Hadamard product and  $W_1,W_2 \in \mathbb{R}^{(100 \times 100)}$ are trainable weights. Next, we take row-wise and column-wise sum of $F_1$ and $F_2$ to obtain two vectors $v_1 \in \mathbb{R}^{(100 \times 1)}$ and $v_2 \in \mathbb{R}^{(100 \times 1)}$:
\begin{equation}
    v_1 = F_1u^T, \;\text{and}\; 
    v_2 = ( uF_2)^T,
\end{equation}
where $u \in \mathbb{R}^{(1 \times 100)}$ is a vector of ones. Although neither $v_1$ or $v_2$ is an exact replica of $\{S_{l,m}\}$, we have observed that empirically, this step  increases the capacity of the network and makes it more robust to random movements of feature patterns. Our intuition for this behaviour is as follows: in practice, there may be other frequency components in $\Omega_{n,l,m}(f) \cup \Omega_{n,l,0}(g)$, other than $\{S_{l,m}\}$, which are invariant to certain pattern movements. Also, most discriminative and robust features may belong to certain frequency bands, and weighted sum of $F_{100 \times 100}$ allows us to give more emphasis to such prominent frequency bands.

\begin{figure*}[!tbp]
\begin{minipage}[t]{0.32\textwidth}
  \centering
    \includegraphics[width=\linewidth]{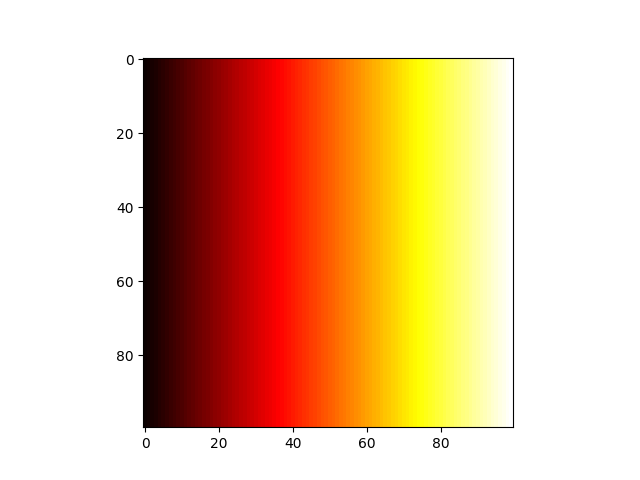}
\label{equ1}
\end{minipage}
\hfill
  \begin{minipage}[t]{0.32\textwidth}
    \centering
          \includegraphics[clip, trim=1.5cm 2.8cm 0.5cm 0.5cm,width=0.68\linewidth]{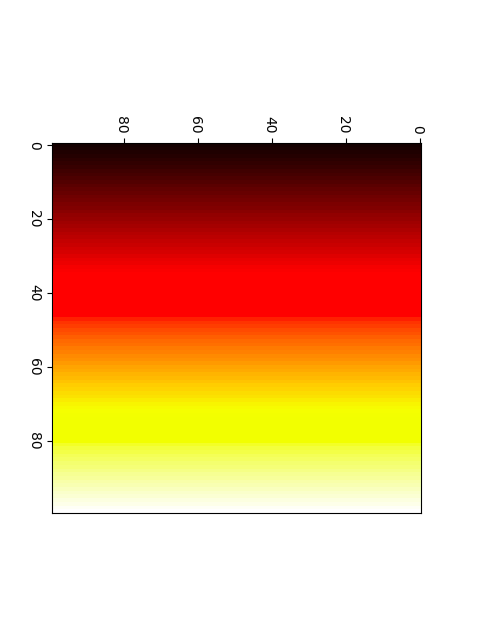}
    \label{equ2}
    \end{minipage}
    \hfill
  \begin{minipage}[t]{0.32\textwidth}
    \centering
          \includegraphics[width=\linewidth]{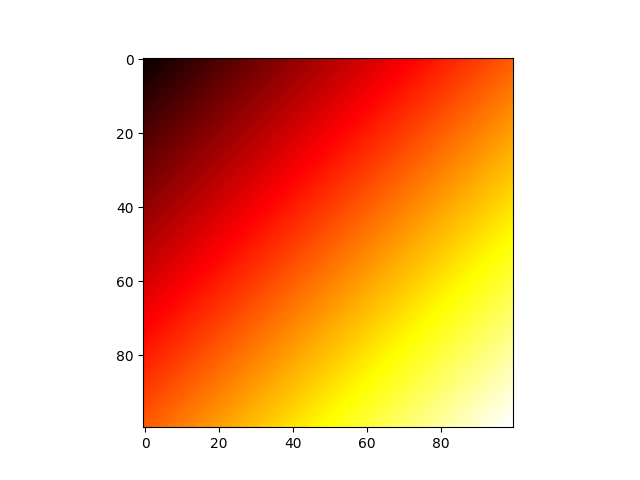}
    \label{recons}
    \end{minipage}
    \caption{The heat-maps of the dense frequency map. \textit{Left:} frequency heat-map with respect to kernel. \textit{Middle:} frequency heat-map with respect to input function. \textit{Right:} frequency The resultant heat-map of the dense frequency map.}
    \label{fig:frequency}
\end{figure*}

\subsection{Experimental Architectures}
\label{sec:architecture}

In this section, we present two experimental architectures to demonstrate the usefulness of volumetric convolution in 3D object recognition tasks. The two types of architectures considered here are with a \emph{single convolution layer} and \emph{multi-convolution layers} respectively. Between these two types, single convolution layer showed better classification performance on popular object datasets with simple 3D shapes, as reported in Section~\ref{sec:ablation}. In contrast, the multi-convolution layer architecture shows better performance for complex 3D shapes, as demonstrated in the same section. 

\subsubsection{Single convolution layer architecture}
\label{sec:single-architecture}
\begin{figure*}[t!]
\centering
\includegraphics[width=1.0\textwidth]{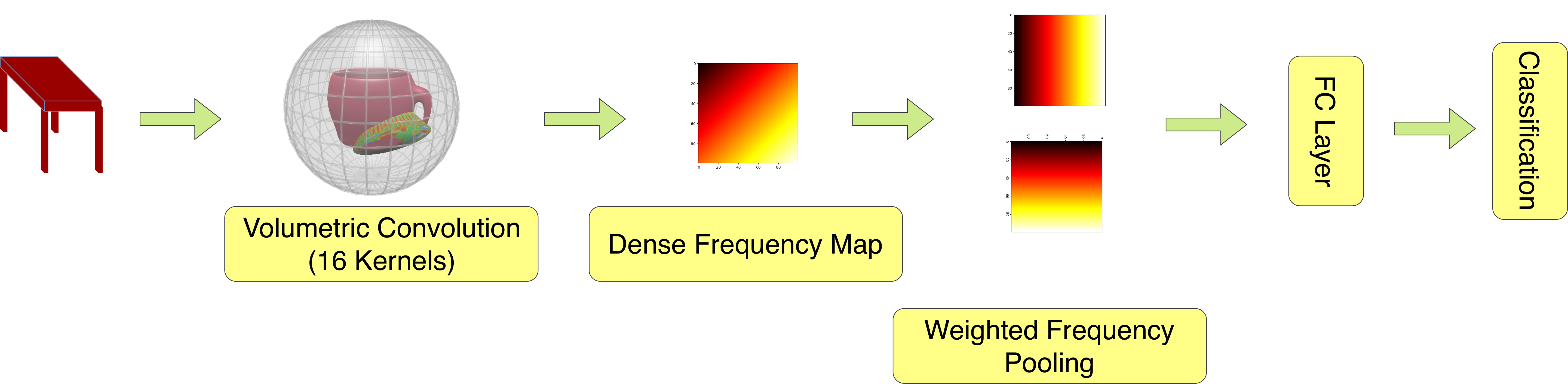}
\caption{The overall experimental architecture.}
\label{fig:archi}
\end{figure*}

{In this architecture,} the object is initially fed to  a volumetric convolution layer with $16$ kernels. Each kernel is translated $10$ times as mentioned in Section~\ref{sec:translation}, which gives a total of $160$ kernels. We use $n=6$ to implement Eq.~\ref{equ:volconvolution}, which gives $100$ dimensional vectors $\Omega_{n,l,m}$ and $\Omega_{n,l,0}$ to represent the input object and each kernel respectively. Convolving input with $16$ kernels results in $160 \times 100 \times 100$ dimensional output feature map, since we take the outer product between $\Omega_{n,l,m}$ and $\Omega_{n,l,0}$ as explained in Section \ref{sec:frequency}. Afterwards, we  perform frequency pooling in two orthogonal directions which reduces the dimensionality of the feature map to $160 \times 100 \times 2$. The output of the frequency pooling layer is then fed to a fully connected layer for classification. We do not use any non-linearity in this single convolution layer architecture. The overall experimental architecture is shown in Fig.~\ref{fig:archi}.

\subsubsection{Multi-convolution layer architecture}

In the multi-convolution layer architecture, the penultimate layer operates similar to the explanation in Section \ref{sec:single-architecture}, while the operation of other (intermediate) convolution layers differs slightly. {The main difference is that} both the input and the output of an  intermediate convolution layer are in spatial domain, as opposed to the  penultimate layer. Let there be $N$ kernels for an intermediate convolution layer. Then, we calculate Zernike moments for both input function and  kernels, and perform convolution as per Eq.~\ref{equ:volconvolution}. From the output, we sample $300$  equi-spaced points for each $\theta$ and $\phi$ direction in the angular space, where $0<\theta<2\pi$ and  $0<\phi<\pi$. To sample points in $r$ direction, we translate each kernel $10$ times by an amount of $0.1$, and perform convolution for each translated state. This overall procedure results in $N$ output feature maps in $\mathbb{B}^3$, where each feature map has $10 \times 300 \times 300$ sampled points in the spatial domain. These feature maps are then fed to a ReLU layer, before being convolved again by the next convolution layer. We apply adaptive frequency pooling only to the penultimate layer, as we do not revert to spatial domain after that.

For both architectures, we use three iterations to calculate the Moore-Penrose pseudo inverse using Eq.~\ref{equ:inverse}. We use a decaying learning rate $lr = 0.1 \times 0.9^{\frac{g_{step}}{3000}}$, where $g_{step}$ is incremented by one per iteration. For training, we use the Adam optimizer with hyper-parameters $\beta_1 = 0.9, \beta_2 = 0.999, \epsilon = 1\times 10^{-8}$. 
All the weights are initialized using a random normal distribution with $0$ mean and $0.5$ standard deviation. All these values are chosen empirically. Since we have decomposed the theoretical derivations into sets of low-cost matrix multiplications, specifically aiming to reduce the computational complexity,  the GPU implementation is highly efficient. For example, the model takes less than 25 minutes for an epoch during the training phase for ModelNet10, with a batch size 2, on a single GTX 1080Ti GPU.

\section{Experiments}\label{sec:experiments}
In this section, we discuss and evaluate the performance of the proposed approach on 3D object recognition and retrieval tasks. We first apply our experimental architecture on five recent datasets, and compare the performance with relevant state-of-the-art works. An extensive ablation study is also reported. We then evaluate the robustness of the captured features against loss of information and finally show that the proposed approach for computing 3D Zernike moments produce richer representations of 3D shapes compared to the conventional approach.


\subsection{Datasets}

\noindent
\textbf{$-$ ModelNet40} 
 contains $40$ object categories and a total of $12{,}311$ CAD models. Train and test sets originally contain $9{,}843$ and $2{,}468$ models respectively. We use the standard train/test split to evaluate our model.

\noindent
\textbf{$-$ ModelNet10}
 is closely related to ModelNet40 dataset, and contains 10 object classes. We use the original train/test split provided by the authors of the dataset, which contains $3991$ models for
training and $908$ models for testing.

\noindent
\textbf{$-$ McGill shape dataset}
 is a benchmark 3D shape dataset with 10 classes: ant, crab, spectacle, hand,
human, octopus, plier, snake, spider and teddy-bear. The dataset contains a total of 255 objects with a variety of pose changes and part articulations.

\noindent
\textbf{$-$ SHREC'17 dataset}
is a challenging state-of-the-art 3D object dataset. This large scale dataset contains about $51{,}300$ 3D models over 55 common categories. Each category is subdivided into several subcategories, but we use only the main 55 categories in our experiments. We use the original split by the authors which is 70\%-30\% for train and test respectively.


\begin{table*}
  \caption{Comparison with state-of-the-art on ModelNet10 (ranked according to performance).}
  \label{table:mod10}
  \centering
  \begin{tabular}{llll}
    \toprule
    \cmidrule(r){1-3}
    Method     & Trainable layers & Trainable Params & ModelNet10   \\
    \midrule
    SO-Net \venue{(CVPR'18)} \cite{li2018so} & 11FC & 60M & 95.7\%     \\
      Kd-Networks \venue{(ICCV'17)} \cite{klokov2017escape}     & 15KD & 4M  & 94.0\%     \\
      \textbf{Ours} & \textbf{(1Conv, 1Adapt. FreqPool, 1FC)} & \textbf{0.7M} & \textbf{93.8\%} \\
         VRN \venue{(NIPS'16)} \cite{brock2016generative}     &  45Conv & 90M & 93.11\%   \\
        Pairwise \venue{(CVPR'16)} \cite{johns2016pairwise}     & 23Conv & 143M & 92.8\%       \\
 
                DeepPano \venue{(SPL'15)} \cite{shi2015deeppano}     & (4Conv, 3FC) & - & 85.45\%     \\
                3DShapeNets \venue{(CVPR'15) } \cite{wu20153d}     & (4-3DConv,2FC) & 38M & 83.5\%  \\ 
                PointNet \venue{(IJCNN'16)} \cite{garcia2016pointnet}  & (2Conv, 2FC) & <1M & 77.6\%    \\

     
    \bottomrule
  \end{tabular}
\end{table*}

\begin{table*}
  \caption{Comparison with state-of-the-art on ModelNet40 (ranked according to performance).}
  \label{table:mod40}
  \centering
  \begin{tabular}{llll}
    \toprule
    \cmidrule(r){1-3}
    Method     & Trainable layers & Trainable Params &  ModelNet40    \\
    \midrule
    SO-Net \venue{(CVPR'18)} \cite{li2018so} & 11FC& 60M & 93.4\%       \\
      Kd-Networks \venue{(ICCV'17)} \cite{klokov2017escape}  &  15KD & 4M & 91.8\%     \\
 \textbf{Ours} & \textbf{(1Conv, 1Adapt. FreqPool, 1FC)} & \textbf{0.7M} & \textbf{91.0\%}\\
         VRN \venue{(NIPS'16)} \cite{brock2016generative}     & 45Conv & 90M & 90.8\%     \\
        Pairwise \venue{(CVPR'16)} \cite{johns2016pairwise} &     23Conv & 143M &  90.7\%      \\
      MVCNN \venue{(ICCV'16)}n \cite{su2015multi}     & (60Conv, 36FC) & 200M & 90.1\%     \\

           PointNet \venue{(CVPR'17)}  \cite{qi2017pointnet} & (5Conv, 2STL)    & 80M & 86.2\%      \\
              ECC \venue{(CVPR'17)} \cite{simonovsky2017dynamic}     & (4Conv, 1FC) & - &   83.2\%    \\
                DeepPano \venue{(SPL'15)} \cite{shi2015deeppano}     & (4Conv, 3FC) & - & 77.63\%     \\
                3DShapeNets \venue{(CVPR'15) } \cite{wu20153d}     & (4-3DConv, 2FC) & 38M & 77\%\\

     
    \bottomrule
  \end{tabular}
\end{table*}

\begin{figure}[t!]
\centering
\includegraphics[width=0.4\textwidth]{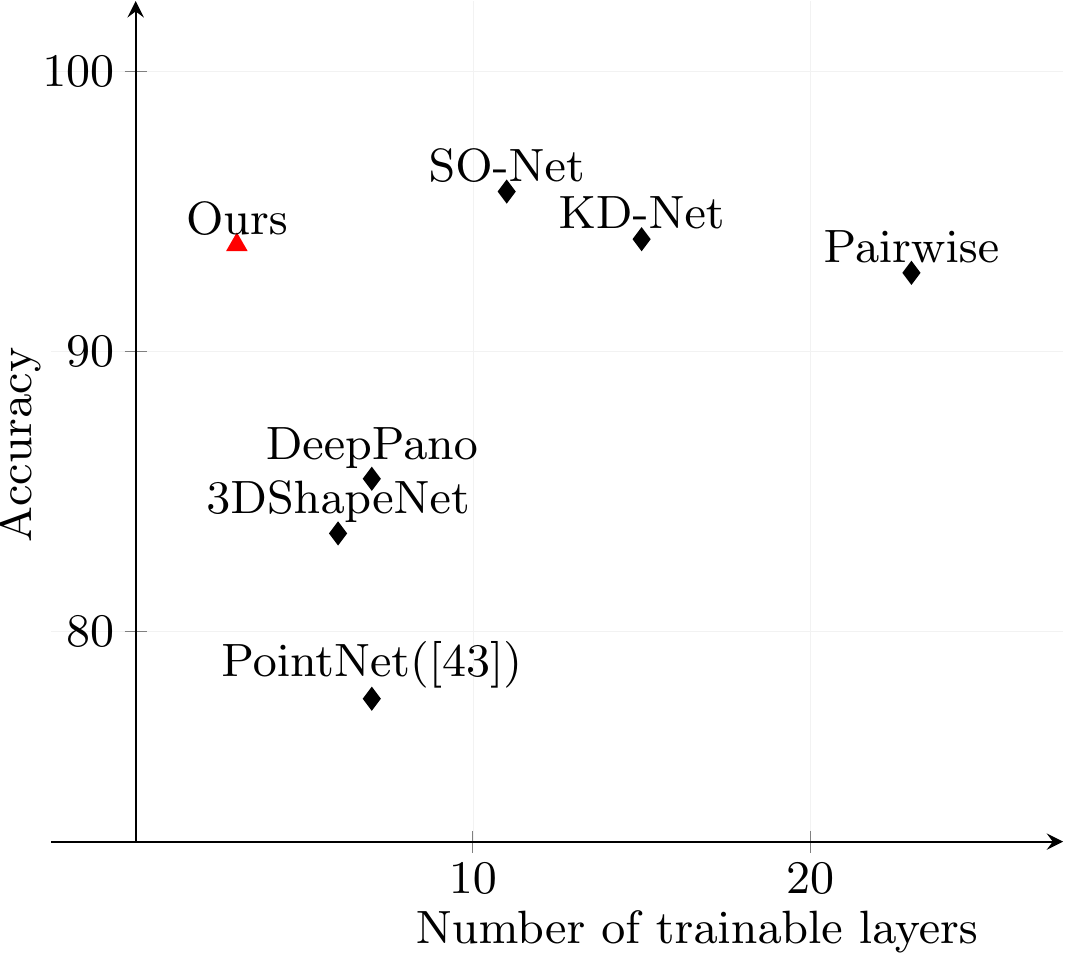}
\caption{Accuracy comparison with state-of-the-art  over ModelNet10 against the number of trainable layers.}
\label{fig:archi10}
\end{figure}

\begin{figure}[t!]
\centering
\includegraphics[width=0.4\textwidth]{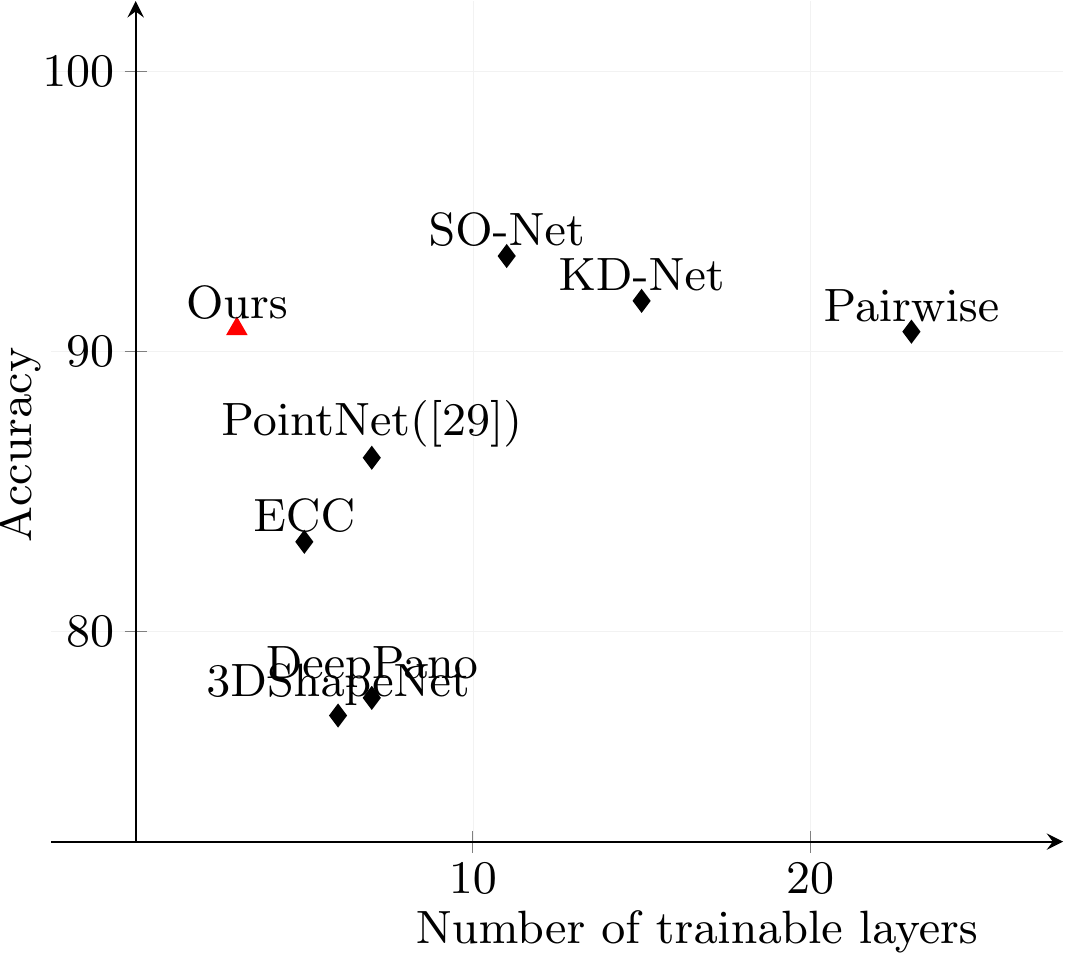}
\caption{Accuracy comparison with state-of-the-art  over ModelNet40 against the number of trainable layers.}
\label{fig:archi40}
\end{figure}

\begin{table}
  \caption{3D object retrieval results comparison with state-of-the-art on McGill Dataset.}
  \label{table:mcgill}
  \centering
  \begin{tabular}{ll}
    \toprule
    \cmidrule(r){1-2}
    Method     & Accuracy  \\
    \midrule
    \cite{tabia2014covariance} & 0.977\% \\
    \cite{agathos2009retrieval} & 0.976\% \\
    \cite{tabia2013compact} & 0.969\% \\
    \cite{papadakis20083d} & 0.957\% \\
    \cite{lavoue2012combination} & 0.925\% \\
    \cite{xie2015deepshape} & 0.988\% \\
    \midrule
    \textbf{Ours} & \textbf{0.988\%} \\
               
     
    \bottomrule
  \end{tabular}
\end{table}

\begin{table}
  \caption{3D object retrieval results comparison with state-of-the-art on SHREC'17.}
  \label{table:shrec}
  \centering
  \begin{tabular}{ll}
    \toprule
    \cmidrule(r){1-2}
    Method     & mAP  \\
    \midrule
\cite{Furuya2016DeepAO} \venue{(BMVC'16)} & 0.476 \\
\cite{esteves2017learning} \venue{(ECCV'18)}  & 0.444 \\
\cite{Tatsuma2009} & 0.418 \\
\cite{bai2016gift} \venue{(CVPR'16)} & 0.406 \\
\midrule
\textbf{Ours} & \textbf{0.452} \\

    \bottomrule
  \end{tabular}
\end{table}

\subsection{3D object classification}
One key feature of our proposed volumetric convolution is that it is a natural extension of planar convolution to spherical domain (specifically $\mathbb{B}^3$). In the same way as a planar kernel finds distributed discriminative patterns across $(x,y)$ plane, volumetric convolution is able to find such patterns distributed across the 3D space. Practically, this should enable our model to capture rich features with less number of layers, compared to other state-of-the-art models, that are somewhat ad-hoc extensions to 3D domain. To demonstrate this, we present the model complexity and accuracy analysis on ModelNet10 and ModelNet40 datasets. Table~\ref{table:mod10} shows the results on ModelNet10. Our model achieves an accuracy of 93.8\% over ModelNet10 with only three trainable layers: one convolution layer, one frequency pooling layer and one fully connected layer. Our accuracy is the third highest, below SO-Net and Kd-Networks. Compared to models such as VRN and PairWise, which have 45 and 23 convolution layers respectively, our model achieves a higher accuracy with a significantly less number of layers. This clearly demonstrates the richness of computed features by volumetric convolution.
Table~\ref{table:mod40} shows the results over ModelNet40. Our model achieves an accuracy of 91.0\% and ranks third, same as the case of ModelNet10. These results demonstrate that our model has a good generalization over a large number of object categories, without losing its advantage as a rich feature computer. Also, our model has only 0.7M trainable parameters, which is a drastically lower  compared to state-of-the-art. This significant reduction in number of parameters is a fair indication of the effectiveness of our \textit{adaptive-frequency-pooling}  layer.

To illustrate the trade-off between model complexity and performance, Fig.~\ref{fig:archi10} and Fig.~\ref{fig:archi40} plot accuracy against number of trainable layers of state-of-the-art models on ModelNet10 and ModelNet40 datasets. These figures clearly show that volumetric convolution is in a better position compared to most recent models, in terms of the trade-off between complexity and accuracy. 
 Note that, our method is not directly comparable with with some other recent works (e.g.,  \cite{kanezaki2016rotationnet,  sedaghat2016orientation, wu2016learning, qi2016volumetric, bai2016gift, maturana2015voxnet}) that use multiple-models and/or data and feature augmentation.

\subsection{3D Object Retrieval}
We evaluate the 3D object retrieval performance of our model on McGill and SHREC'17 datasets. We obtain the 200-dimensional feature descriptor after the frequency pooling layer, and measure the cosine similarity between the query shape and the shapes in the database. We first train the model as a classifier using train set, with softmax cross entropy as the loss function, and then use test set to evaluate the retrieval performance. The results for McGill dataset are shown in Table~\ref{table:mcgill}. We use the nearest neighbour performance measure for this task. For McGill dataset, we compare the performance of our model with six state-of-the-art techniques:\cite{tabia2014covariance},
   \cite{agathos2009retrieval}, 
    \cite{tabia2013compact}, 
    \cite{papadakis20083d}, 
   \cite{lavoue2012combination} and  
    \cite{xie2015deepshape}. As shown in Table~\ref{table:mcgill}, our feature vector is able to match the state of the art results achieved by \cite{xie2015deepshape}.
    
Table~\ref{table:shrec} depicts the performance comparison on SHREC'17 dataset (as reported in \cite{esteves2017learning}). This dataset includes  random
$\mathbb{SO}(3)$ perturbations. We use mean average precision (mAP) to evaluate the performance as done in other state-of-the-art techniques. Our model achieves the second best performance with a mAP value of 0.452, which is only a small drop (0.024) compared to \cite{Furuya2016DeepAO}. Overall, the results for 3D object retrieval task clearly demonstrate the richness of our proposed feature descriptors.

\subsection{Ablation Study}
\label{sec:ablation}
To justify our model choices, we perform an extensive ablation study on ModelNet10 and SHREC'17 datasets. The results are reported in Table~\ref{table:ablation}. First, we use two and three convolution layers instead of one. Accuracy drops from 93.8\% to 92.0\% and 89.8\% in the cases of two layers and three layers respectively. This is an interesting result, as usually one might expect the model to compute richer features as the number of layers increase. However, it is known that in deep models, the accuracy does not always increase with the number of layer due to factors such as over-fitting on training set.
Since our main focus of this work is to provide the theoretical framework of volumetric convolution and implement it as a differentiable layer that can be integrated into any deep architecture, we do not extensively investigate other architectural choices or regularization measures that might perform well with multiple layers. Rather, our main focus of the experiments is to show the richness of features computed using volumetric convolution. 

Then, we replace the volumetric convolution layer with a spherical convolution layer, and achieve an accuracy of 76.2\%. This is perhaps one of the most important comparisons in our ablation study. This clearly shows that modeling a 3D object and convolving it in $\mathbb{B}^3$ gives superior results as opposed to spherical convolution, which performs convolution in $\mathbb{S}^2$. To demonstrate one practical use-case of axial symmetry measurement of functions in $\mathbb{B}^3$, we measure the symmetry of objects around four equi-angular axes, and concatenate these measurement values to form a feature vector. Then we feed the generated feature vector to a fully connected layer to classify objects. Since we theoretically derive and implement the axial symmetry measurement formula as a fully differentiable module, this setting can be trained via backpropagation. We were able to achieve a 60.4\% accuracy over ModelNet10 from using this simple hand-crafted feature.

Next, we investigate the effect of using various pooling mechanisms, instead of the proposed adaptive frequency pooling. In all the pooling operations, we create a $(100 \times 100)$ dimensional dense frequency map and perform pooling both row-wise and column-wise. First, we pool the outputs of the 16 kernels using mean-pooling and max-pooling, which drops the accuracy below 90\% in both cases. Then we concatenate the kernel outputs and directly feed it to the fully connected layer, and achieve an accuracy of  87.8\%. \cite{ilse2018attention} recently proposed two novel attention based multiple instance learning (MIL) pooling mechanisms, that has trainable weights. Since our intuition for using frequency pooling is to capture prominent frequency bands from frequency maps, we test the model using aforementioned MIL pooling mechanisms as it can learn to give attention to different frequency bands. We first construct the  dense frequency map and then apply the two MIL pooling mechanisms---gated and non-gated---to achieve 89.8\% and 90.3\% accuracy respectively.

Recently, \cite{liu2017deep} introduced a novel learning framework that gives angular representations on hyperspheres. This framework is supervised by two novel loss formulations that utilizes the angular similarity between the final descriptors. Since it is fair to assume that angular similarity plays a significant role in our model too---specially due to volumetric convolution's equivariance to 3D rotation group---we test the performance of weighted-softmax function and generalized-angular-softmax function proposed by \cite{liu2017deep} in our experiments. However, as illustrated in Table~\ref{table:ablation}, neither of these loss functions are able to outperform softmax loss function in our setting. 

Furthermore, we use different similarity measures for retrieving 3D objects and compare the performances. As shown in Table~\ref{table:ablation}, cosine similarity performs best while Euclidean, KL and Bhattacharya give inferior results.

\begin{table}
   \caption{Ablation study of the proposed architecture on ModelNet10 and SHREC'17 datasets. Here, ``$+$'' sign refers to ``with'' and ``$-$'' sign refers to ``without''.}
  \label{table:ablation}
   \centering
  \begin{tabular}{l c}    
    \toprule
    \cmidrule(r){1-2}
 \textbf{3D Object Classification} & \\
 \midrule
    Method     & Accuracy     \\
    \midrule
    Final Architecture (FA)  &     93.8\%  \\
    FA (2Conv) &     92.0\%  \\
    FA (3Conv) &     89.8\%  \\
    FA $-$ VolCNN $+$ SphCNN      &  76.2\%      \\
    Axial symmetry features & 60.4\% \\
    \midrule
    FA $-$ Adapt. FreqPool $+$  MeanPool   & 84.2\%     \\
    FA $-$ Adapt. FreqPool $+$  MaxPool   & 86.7\%     \\
   FA $-$ Adapt. FreqPool $+$  FeatureConcat   & 87.8\% \\
    FA $-$ Adapt. FreqPool $+$  MILAPooling~\citep{ilse2018attention}  & 90.3\% \\
    FA $-$ Adapt. FreqPool $+$  MILGAPooling~\citep{ilse2018attention}   & 89.8\% \\
    \midrule
    FA $+$ WSoftmax \citep{liu2017deep} &  92.8\%\\
    FA $+$ GASoftmax \citep{liu2017deep}  & 90.7\% \\
    \toprule
    \textbf{3D Object Retrieval} & \\
    \midrule
    Method & mAP \\
    \midrule
    FA (Cosine Similarity) & 0.452 \\
    FA (Euclidean Distance) & 0.386 \\
    FA (KL Divergence) & 0.320 \\
    FA (Bhattacharyya Distance) & 0.354 \\
    \bottomrule
  \end{tabular}
\end{table}

\subsection{Classification of highly non-polar and textured objects}
The ablation study shown in Table \ref{tab:multilayer} 
depicts that accuracy drops when multi-layer architectures are used in object classification.  In this section, we explore a possible reason for this behaviour.

Two key features of our convolution layer are: (a) the ability to jointly model both shape and texture information, and (b) handling non-polar (i.e. dense in $\mathbb{B}^3$) objects. However, the dataset used for the ablation study experiment---ModelNet10---contains relatively simpler shapes with uniform texture. Therefore, using more layers (thus more parameters) can cause overfitting on the training set, as our network is able to capture highly discriminative features using a single volumetric convolution layer, which can cause a drop in test accuracy. To verify this, we employ a multi-layer architecture to classify a more challenging dataset, where objects are highly non-polar and textured.

To this end, we  sample $1000$ 3D brain scan images from the large-scale OASIS-3 dataset (Fotenos et al. 2005). 
OASIS-3 is a compilation of 3D MRI brain scans obtained from over $1000$ participants, collected over the course of 30 years. Participants include $609$ cognitively normal adults and $489$ individuals at various stages of Alzheimer’s Disease aged between 42-95yrs. We split the sampled data in to train and test sets, comprising of $800$ and $200$ scans respectively. In both the train and test sets, we include equal numbers of Alzheimer and normal cases to avoid any bias.  Afterwards, we train and test our network on the sampled data with softmax cross entropy as the loss function. Table \ref{tab:multilayer} 
shows the performance against the number of layers used in the network.

\begin{table}
\caption{Performance of multi-layer architectures for highly non-polar and textured shape classification. Our model shows an improvement with higher number of layers.}
\centering
\label{tab:multilayer}
 \begin{tabular}{  c  c   }
     \toprule
    \cmidrule(r){1-2}
 \textbf{Model} & \textbf{Accuracy}  \\ 
\midrule
Ours (1 Conv layer) & 69.4\% \\ 
Ours (2 Conv layers) & 78.8\%\\
Ours (3 Conv layers) &  \textbf{83.2\%}\\
Ours (4 Conv layers) &  82.8\%\\
\bottomrule
  \end{tabular}
  \end{table}
  
In this experiment, we used 16 convolution kernels in each layer. As  evident from Table \ref{tab:multilayer}, increasing the number of convolution layers improve the classification accuracy, up to three layers. Thus, it can be concluded that dense objects with texture allow our network to showcase its full capacity.

\subsection{Equivariance to local pattern movements}\label{sec:local_movements}

For the translation and rotation of local feature patterns, which results in non-rigid deformations of the global shape, our proposed convolution operator ensures equivariance. In this experiment, we evaluate the robustness of our proposed network to such movements of local feature patterns. To this end, we radially translate and rotate local feature patterns in 3D and compare the behaviour of our approach with a traditional spatial-domain convolution method (\cite{maturana2015voxnet}). Furthermore, we also test our approach with an even more difficult set of `random' movements of local patterns.

Initially, we get the heat kernel signature of the 3D shape with 20 eigen vectors and a time stamp of 10. Afterwards, we get the vertices associated with the highest 10\% of the heat response and cluster them using DBSCAN algorithm \cite{ester1996density}. We then find the centroid of each cluster, and get the 50 closest sample points to each centroid to obtain a set of sample point clusters. We then move each cluster independently and classify the final set of points using networks already trained on ModelNet10. The results are shown in Table~\ref{tab:lmovement}.

\begin{table*}
  \centering
    \captionof{table}{Performance comparison on local object-part movement resulting in global non-rigid deformations. Accuracies are reported for the ModelNet10 dataset. Performance drop under global deformations is shown in \textcolor{blue}{blue}. Our approach demonstrates minimal performance drop under totally random deformations which signifies the strong invariance behaviour of proposed approach.} 
   \label{tab:lmovement}
  


    \begin{tabular}{ | c | c | c | c | }
  \hline
 & \textbf{Original Shape} & \textbf{Rot + Radial trans} & \textbf{Random} \\ 
 \hline
 \textbf{Ours} & 93.8\% & 91.4\% {\textcolor{blue}{($\downarrow$2.4)}}  & 88.5\% {\textcolor{blue}{($\downarrow$5.3)}} \\
 \textbf{Ours without FreqPool} & 82.8\% & 78.3\% {\textcolor{blue}{($\downarrow$4.5)}} & 61.4\% {\textcolor{blue}{($\downarrow$21.4)}} \\
 \textbf{VoxNet} & 90.4\% & 43.8\% {\textcolor{blue}{($\downarrow$46.6)}} & 42.1\% {\textcolor{blue}{($\downarrow$48.3)}} \\
 \hline
 
  \end{tabular}
\end{table*}

As illustrated in Table \ref{tab:lmovement}, our network is robust to both random and rotational + translational movements of local patterns. Furthermore, when we remove the frequency pooling after the convolution layer, and connect the fully connected layer directly to the response of the convolution, the network becomes less robust to random movements of local patterns. Overall, even with highly challenging severe deformations, we note that the proposed approach does not experience a significant drop in the accuracy, compared to the spatial-domain convolution based approach. This behaviour signifies the strong invariance capability of proposed convolution operator.

\subsection{Robustness against information loss}
\begin{figure}[!tbp]
\begin{minipage}[t]{0.49\textwidth}
  \centering
    \includegraphics[width=0.7\textwidth]{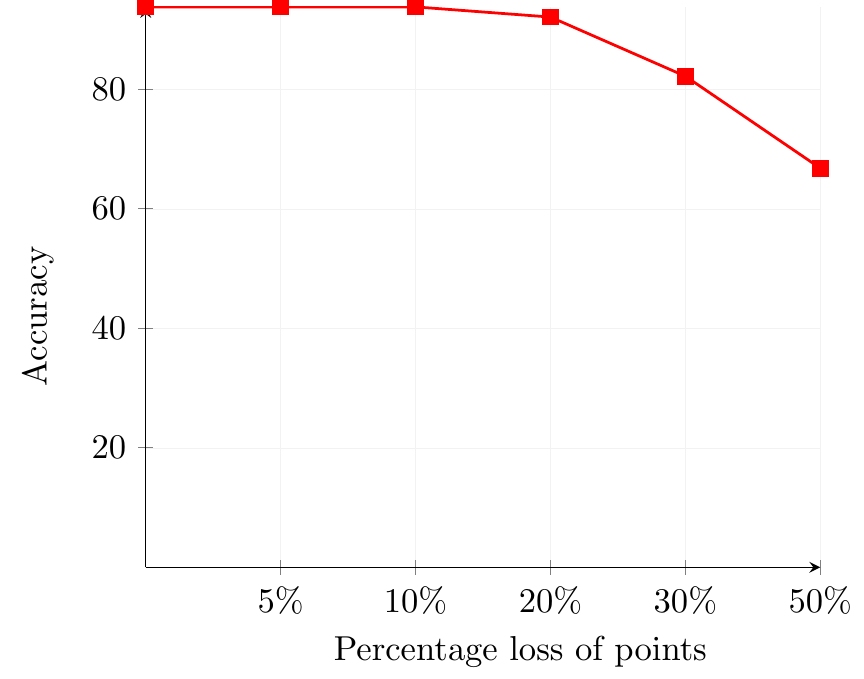}
    \captionof{figure}{The robustness of the proposed model against missing data. The accuracy drop is less than $30\%$ at a high data loss rate of $50\%$.}
\label{robust}
\end{minipage}
\hfill
  \begin{minipage}[t]{0.49\textwidth}
    \centering
          \includegraphics[width=0.7\textwidth]{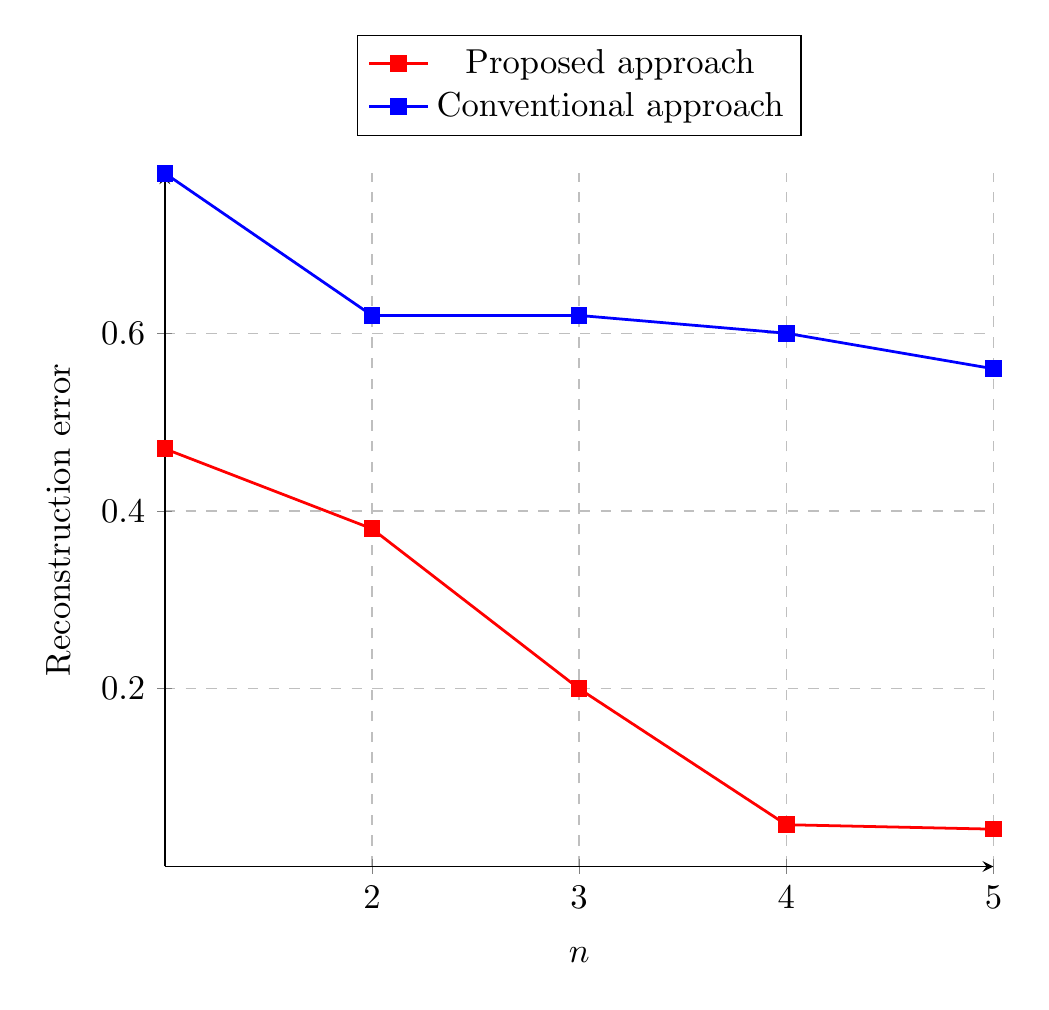}
    \captionof{figure}{The mean reconstruction error Vs `$n$'. Our Zernike frequencies computation approach has far less error than the conventional approach.}
    \label{recons}
    \end{minipage}
\end{figure}

One critical requirement of a 3D object classification model is to be robust against information loss. To demonstrate the effectiveness of our proposed features in this aspect, we randomly remove data points from the objects in validation set, and evaluate model performance. The results are illustrated in Fig.~\ref{robust}. The model shows no performance loss until $20\%$ of the data points are lost, and only gradually drops to an accuracy level of $66.8$ at a $50\%$ data loss. This implies that the proposed model is robust against data loss and can work well for incomplete shapes.

\subsection{Approximation Accuracy of 3D Zernike moments calculation approach}
In Sec.~\ref{shapemodeling}, we proposed an alternative method to calculate 3D Zernike moments (Eq.~\ref{reconstruction2}, \ref{linear}), instead of the conventional approach (Eq.~\ref{omega}).  We hypothesized that moments obtained using the former has a closer resemblance to the original shape, due to the impact of finite number of frequency terms. In this section, we demonstrate the validity of our hypothesis through experiments. To this end, we compute moments for the shapes in the validation set of ModelNet10 dataset using both approaches, and compare the mean reconstruction error defined as: $\frac{1}{T}\sum_t^T\norm{f(t)-\sum_{n}\sum_{l}\sum_{m}\Omega_{n,l,m}Z_{n,l,m}(t)}$, where $T$ is the total number of points and $t \in \mathbb{B}^3$. Fig.~ \ref{recons} shows the results. In both approaches, the mean reconstruction error decreases as $n$ increases. However, our approach shows a significantly low mean reconstruction error of $0.0467\%$ at $n=6$ compared to the conventional approach, which has a mean reconstruction error of $0.56\%$ at same $n$. This result also justifies the utility of Zernike moments for modeling complex 3D shapes.

\section{Comparison with Invariant Approaches}
In the literature, the 3D shape analysis techniques with invariance properties have been proposed for both continuous surfaces and discrete point clouds. For the former representations, a Riemannian metric was proposed for parameterized 3D surfaces that is invariant to shape re-parametrizations \citep{kurtek2010novel}.  In our case, invariance to re-parametrization group is less practical as we are working with discretized 3D shapes. Further, incorporating such distance metrics and parametrizations of real-world 3D shapes within deep feature learning models is  a fairly challenging and largely unsolved problem. For the case of point clouds, permutation invariance has been studied for deep convolutional \citep{qi2017pointnet} and graph convolutional networks in \citep{maron2018invariant,wang2018dynamic}. The above works show that achieving permutation invariance is relatively simple in deep architectures.

Graph based approaches have been proposed to work on non-Euclidean topologies and are thus suitable to operate on 3D surfaces \citep{bronstein2017geometric}. An input surface is converted to a graphical representation (e.g., polygon mesh) and converted to spectral domain where convolution is performed \citep{bruna2013spectral,defferrard2016convolutional, boscaini2015learning, henaff2015deep}. A different set of methods first reduce the complexity of input data by projecting them in a parametric 2D representation space and then apply convolutions to learn features. Finally, \citep{masci2015geodesic, monti2017geometric, boscaini2016learning} perform convolution within local surface patches and thus provide invariance to surface deformations. However, the desirable invariance to deformations is generally dictated by the end-task and may not always be desirable since significant deformations can change object functionality, affordance and semantics \citep{su2018splatnet}. As we explain in the next paragraph, our fully learnable network allows task-dependent learning of invariance to shape deformations. Further, all of the above approaches learn representations on the shape surface or its 2D transformed version and do not consider the volumetric nature of 3D shapes.  

In comparison to above mentioned approaches, we propose a novel convolutional operator in $\mathbb{B}^3$ that is suitable for volumetric shapes. Our proposed convolution operator is equivariant to isometric transformations (rotation, translation), and is also robust to non-isometric variations such as deformations (radial translations of local object parts) and shape articulations (scaling and local part rotations) (as shown in Sec.~\ref{sec:local_movements}). Deep learning based solutions that can achieve invariance to all types of deformations are relatively less explored\footnote{we refer the reader to \citep{cohen2018general} for an excellent review on group equivariant CNNs.} and, to the best of our knowledge, ours is the first roto-translation equivariant convolution operator inside the unit-ball. The advantage of our approach over graph based invariant models is the ability to learn representations on volumetric shapes. As an example, a recent work \citep{maron2018invariant} investigates invariance and equivariance for graph networks but only considering the linear layers (not the convolution ones) in the 2D case. The extension of our proposed convolution operator to arbitrary graphs and all possible deformations is an interesting research problem but beyond the scope of current work. 

It is noteworthy that the end-to-end representation learning in our case automatically enforces invariance to deformations and articulations depending on the end-task. In comparison, the traditional approaches \citep{carriere2015stable, reininghaus2015stable} for topological data analysis propose hand-crafted descriptors (based on persistence diagrams) that are invariant to only certain classes of deformations (intrinsic and extrinsic isometries). As a result, these descriptors are relatively less generalizable and their manual design offers less flexibility for new problems. We have conducted an experiment in this regard, where ModelNet10 shapes are first deformed by random movements of local object parts and their feature representations are used for final classification (Sec.~\ref{sec:local_movements}). We achieve a classification accuracy quite close to that of original shapes, showing that the deformed and articulated shapes are mapped close to the original unaltered 3D shapes in the learned feature space. 

\section{Conclusion}\label{sec:conclusion}
In this work, we derive a novel `\emph{volumetric convolution}' using 3D Zernike polynomials, which can learn feature representations in $\mathbb{B}^3$. We develop the underlying theoretical foundations for volumetric convolution and demonstrate how it can be efficiently computed and implemented using low-cost matrix multiplications. Furthermore, we propose a novel, fully differentiable method to measure the axial symmetry of a function in $\mathbb{B}^3$ around an arbitrary axis, using 3D Zernike polynomials and demonstrate one possible use case by proposing a simple hand-crafted descriptor. Finally, using volumetric convolution as a building tool, we propose an experimental architecture, that gives competitive results over state-of-the-art with a relatively shallow network, in 3D object recognition and retrieval tasks. In this experimental architecture, we introduce a novel frequency pooling layer, which can learn frequency bands in which the most discriminative features lie. One drawback of the current volumetric convolution operator is that 3D Zernike polynomials loose its orthogonality when a 3D translation is applied. This prevents volumetric convolution from achieving automatic translation invariance. Therefore one immediate extension to this work would be to investigate novel orthogonal and complete basis polynomials in a unit ball, which preserves its orthogonality when translated. Such polynomials would make it possible to achieve translation invariance more efficiently---compared to the proposed method---as then, the conversion from spatial domain to spectral domain at each translation of the kernel is not necessary. 


\bibliographystyle{spbasic}      
\bibliography{egbib1}   


\end{document}